\documentclass[lettersize,journal]{IEEEtran}
\usepackage{amsmath,amsfonts}
\usepackage{array}
\usepackage[caption=false,font=normalsize,labelfont=sf,textfont=sf]{subfig}
\usepackage{textcomp}
\usepackage{stfloats}
\usepackage{url}
\usepackage{verbatim}
\usepackage{graphicx}
\usepackage{cite}
\usepackage{array}
\newcolumntype{P}[1]{>{\centering\arraybackslash}p{#1}}
\usepackage{booktabs}
\usepackage{caption}
\usepackage{multirow}
\usepackage{epigraph}
\usepackage{bbm}
\usepackage[ruled,linesnumbered]{algorithm2e}
\usepackage{algpseudocode}
\usepackage[colorlinks,
linkcolor=blue,
anchorcolor=black,
citecolor=black]{hyperref}

\begin{document}
\bstctlcite{IEEEexample:BSTcontrol}
\title{COutfitGAN: Learning to Synthesize Compatible Outfits Supervised by Silhouette Masks \\and Fashion Styles} 

\author{Dongliang Zhou, Haijun Zhang, Qun Li, Jianghong Ma, and Xiaofei Xu
\thanks{This work was supported in part by the National Natural Science Foundation of China under Grant no. 61972112 and no. 61832004, the Guangdong Basic and Applied Basic Research Foundation under Grant no. 2021B1515020088, the Shenzhen Science and Technology Program under Grant no. JCYJ20210324131203009, and the HITSZ-J\&A Joint Laboratory of Digital Design and Intelligent Fabrication under Grant no. HITSZ-J\&A-2021A01.}
\thanks{D. Zhou, H. Zhang, Q. Li, J. Ma and X. Xu are with the Department of Computer Science, Harbin Institute of Technology, Shenzhen, 518055 China. Corresponding author: Haijun Zhang, e-mail: hjzhang@hit.edu.cn.}}

\markboth{Journal of \LaTeX\ Class Files,~Vol.~14, No.~8, August~2021}%
{Shell \MakeLowercase{\textit{et al.}}: A Sample Article Using IEEEtran.cls for IEEE Journals}


\maketitle

\begin{abstract}
How to recommend outfits has gained considerable attention in both academia and industry in recent years.
Many studies have been carried out regarding fashion compatibility learning, to determine whether the fashion items in an outfit are compatible or not.
These methods mainly focus on evaluating the compatibility of existing outfits and rarely consider applying such knowledge to `design' new fashion items.
We propose the new task of generating complementary and compatible fashion items based on an arbitrary number of given fashion items.
In particular, given some fashion items that can make up an outfit, the aim of this paper is to synthesize photo-realistic images of other, complementary, fashion items that are compatible with the given ones.
To achieve this, we propose an outfit generation framework, referred to as COutfitGAN, which includes a pyramid style extractor, an outfit generator, a UNet-based real/fake discriminator, and a collocation discriminator.
To train and evaluate this framework, we collected a large-scale fashion outfit dataset with over 200K outfits and 800K fashion items from the Internet. Extensive experiments show that COutfitGAN outperforms other baselines in terms of similarity, authenticity, and compatibility measurements.
\end{abstract}

\begin{IEEEkeywords}
Fashion synthesis, compatibility learning, image-to-image translation, fashion analysis, generative adversarial network.
\end{IEEEkeywords}

\section{Introduction}

\IEEEPARstart{R}{ecommending} outfits has become increasingly popular in online fashion shops due to their powerful recommendation ability from a fashion compatibility perspective rather than relying on just a single fashion item itself.
Coco Chanel succinctly stated its concept when she said ``Fashion is architecture: it is a matter of proportions.'' An outfit, which contains a set of fashion items, is expected to be compatible from a visual perspective.
The aim of recommending an outfit is to suggest a set of fashion items to a customer that are composable and compatible.
Many applications, such as WEAR\footnote{\url{https://wear.jp/}}, iFashion\footnote{\url{https://if.taobao.com/}}, and Xiaohongshu\footnote{\url{https://www.xiaohongshu.com/}}, have launched outfit recommendation systems for e-commerce customers.
On the other hand, outfit recommendation is also gaining increasing attention in the fields of computer vision\cite{mcauley2015image, vasileva2018learning,veit2015learning,liu2019toward,liu2019collocating}, recommender systems\cite{cui2019dressing, li2020hierarchical,yang2021attribute}, and multimedia communities\cite{han2017learning, feng2019interpretable,li2017mining,ding2021modeling,zhan2021a3,wang2019diagnosis}.

\begin{figure}[t]
	\centering
	\includegraphics[width=0.49\textwidth, height=0.338194\textwidth]{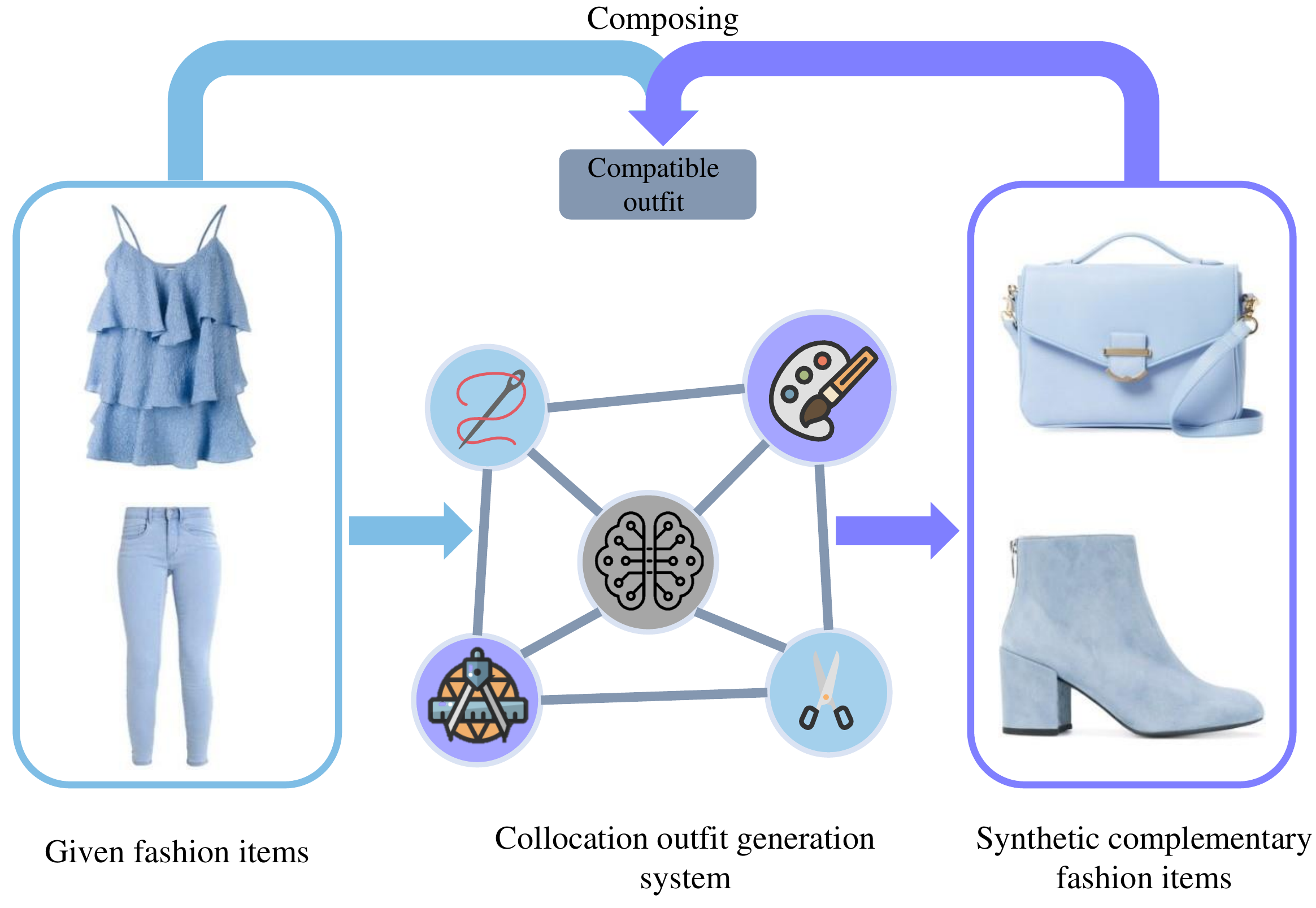}
	\caption{Generating outfits compatible with arbitrary fashion items.}
	\label{cover}
	\vspace{-0.5cm}
\end{figure}

The approaches for outfit recommendation can be divided into three types: determining whether an outfit is compatible or not\cite{mcauley2015image,veit2015learning,vasileva2018learning,han2017learning,cui2019dressing,li2020hierarchical}, determining why an outfit is compatible or not\cite{wang2019diagnosis, yang2021attribute}, and how to create a compatible outfit\cite{liu2019toward,liu2019collocating,yu2019personalized}.
To determine whether an outfit is compatible or not, McAuley \textit{et al.} \cite{mcauley2015image} first used a pre-trained convolutional neural network (CNN) to extract the feature embeddings of each fashion item, and then they computed the distance between each pair of fashion items in their common feature embedding space.
In later studies, Veit \textit{et al.} \cite{veit2015learning} used a Siamese network and Vasileva \textit{et al.} \cite{vasileva2018learning} used a category-based mapping network to find better embedding spaces.
Unlike these metric learning-based methods, Han \textit{et al.} \cite{han2017learning} modeled the fashion compatibility problem as a sequence.
They assumed that a human judges outfit compatibility through bi-directional scanning, from up to down or in the opposite direction.
Another mainstream idea that has emerged to treat compatibility is to use graph neural networks (GNNs).
For example, Cui \textit{et al.} \cite{cui2019dressing} employed a node-wise GNN, and Li \textit{et al.} \cite{li2020hierarchical} adopted a hierarchical GNN, to improve the recommendations of an outfit.
To determine why an outfit is compatible or not, Wang \textit{et al.} \cite{wang2019diagnosis} proposed a diagnosis module using gradients to analyze the rationale behind the compatibility of an outfit.
Subsequently, Yang \textit{et al.}\cite{yang2021attribute} employed the attribute information of fashion items to explain their compatibility with an attention network.
added{However,} there has been very little research on learning how to create a compatible outfit.
Liu \textit{et al.} \cite{liu2019toward} first proposed an Attribute-GAN model to synthesize complementary fashion items in the upper and lower clothing domains.
In a later study, Liu \textit{et al.} \cite{liu2019collocating} further improved the quality of the synthesis of compatible fashion items, employing a multi-discriminator architecture.
Yu \textit{et al.} \cite{yu2019personalized} proposed a synthesis framework to generate complementary fashion items taking into account their compatibility and personalization.
However, these methods only focus on generating a single complementary fashion item.
It has been rare to consider the generation of a whole outfit, conditioning on uncertain fashion items.

To address these problems, we have worked on a novel task: to recommend, in a generative way, an outfit, i.e., outfit generation.
As shown in Fig.~\ref{cover}, outfit generation aims at putting together complementary fashion items, when given arbitrary fashion items, to compose a compatible outfit.
In essence, outfit generation can be regarded as a special image-to-image translation task.
General image-to-image translation includes supervised and unsupervised methods.
In particular, supervised image-to-image translation methods, e.g., Pix2Pix\cite{pix2pix2017} and Pix2PixHD\cite{wang2018pix2pixHD}, translate images from one domain to another domain using ground-truth images as supervision information.
On the other hand, unsupervised methods, e.g., CycleGAN\cite{CycleGAN2017}, MUNIT\cite{huang2018munit}, and CUT\cite{park2020cut}, translate images from one domain to another domain without ground truths.
These methods specialize in processing input images that are spatially aligned to output images by leveraging the encoder--decoder \cite{wu2016google} or UNet\cite{long2015fully} architectures.
However, not all of the cross-domain image data are spatially well aligned in practice.
For example, for an outfit, its fashion items may contain diverse silhouettes, and usually have no apparent spatial alignment with each other.
Furthermore, general image-to-image translations take one image as input and produce another image as output, one at a time.
However, the number of input images in outfit generation may be uncertain.
Compared with aligned pairwise image translation, the generation of fashion items is more challenging, due to their diversity, the variable spatial correspondence between input and output images, and the uncertain number of given fashion items.
To tackle these issues, we add silhouette information as supervision information for the generation of complementary items, to reduce the difficulty of the mapping.
This information, which has also been used for other fashion synthesis tasks \cite{han2018viton,zhu2019progressive,yan2022toward}, can provide explicit spatial supervision for generating targeted fashion items during the training phase.
It is worth pointing out that during the inference phase, silhouette information can be provided by the users for better controllability or selected randomly from a given silhouette database to generate diversified items.
For modeling fashion item styles from a variable number of given images, we also attempt to learn the styles of the target fashion items from the given ones with a pyramid style extractor.
The silhouette information and fashion item styles are then progressively fused with our proposed silhouette and style fusion blocks.
More specifically, this paper proposes a compatible outfit generation framework (COutfitGAN) that starts from arbitrarily given fashion items.
In COutfitGAN, a pyramid style extractor is used to encode the fashion styles of the given fashion items.
Then we use a novel outfit generator, which progressively includes several silhouette and style fusion blocks according to the developed pyramid style extractor.
Furthermore, to make the synthesized images more photo-realistic, we employ a UNet-based discriminator \cite{schonfeld2020u} to supervise the generation process, to improve the authenticity.
We have also designed a collocation discriminator to supervise the synthesized fashion items so that they will be compatible with the given fashion items from a contrastive learning \cite{chopra2005learning} perspective.
In order to examine the performance of our proposed framework, we built a large-scale dataset, called OutfitSet, which includes 200K outfits associated with 800K fashion items crawled from Polyvore\footnote{\url{http://www.polyvore.com/}}.
The fashion items in OutfitSet contain rich categorical and silhouette information.
Our key contributions can be summarized as follows:

\begin{enumerate}
	\item To the best of our knowledge, we are the first to propose an outfit generation framework to synthesize a whole outfit given an arbitrary number of fashion items.
	\item We propose a pyramid style extraction module, which models an arbitrary number of given fashion items and predicts the fashion styles of complementary items.
We also propose a novel outfit generator that includes several fusion blocks aiming at synthesizing the silhouette and style into a new fashion item.
	\item We employ a UNet-based discriminator to improve the synthesis in terms of visual authenticity.
We also propose a collocation discriminator, which can supervise the fashion compatibility among the fashion items from a local to a global perspective, by contrastive learning.
\end{enumerate}
 
The rest of this paper is organized as follows.
Section \ref{related_work} reviews the related literature.
In Section \ref{method}, we describe the proposed outfit generation framework.
In Section \ref{experiment}, the qualitative and quantitative results on our constructed dataset are reported and discussed.
Section \ref{cnc} concludes this paper and discusses possible future work.
\section{Related Work}
\label{related_work}
Our research falls into the fields of outfit recommendation and image-to-image translation, which have been examined in a large body of literature.
In this section, we briefly review the related work in these two fields.
We will also highlight the features of our work in comparison to the relevant previous studies.

\textbf{Outfit Recommendation}.
In general, outfit recommendation is a highly subjective task in fashion analysis and recommendation.
Previous work \cite{mcauley2015image, vasileva2018learning,veit2015learning,cui2019dressing, li2020hierarchical,han2017learning, feng2019interpretable,li2017mining,ding2021modeling,zhan2021a3} mainly focused on finding whether given outfits are compatible or not, in a discriminative way.
McAuley \textit{et al.} \cite{mcauley2015image} first projected fashion items from visual space into a common embedding space with a pre-trained CNN.
They then compared the distance of each pair of fashion items with metric learning.
In particular, they assumed that a smaller distance indicates that the outfit is more compatible.
On the basis of the above metric learning, Veit \textit{et al.} \cite{veit2015learning} employed a learnable CNN to extract feature embedding under a Siamese structure.
Unlike the above metric learning-based methods, Han \textit{et al.} \cite{han2017learning} modeled the fashion compatibility problem as a sequence, and they assumed that humans determine the compatibility of an outfit using a bi-directional scanning, from up to down or in the opposite direction.
They also proposed fashion compatibility classification and fill-in-the-blank (FITB) as basic metrics to evaluate the effectiveness of their model.
On the other hand, several studies using GNNs have emerged to address compatibility modeling.
For example, researchers have employed a node-wise GNN\cite{cui2019dressing} and a hierarchical GNN\cite{li2020hierarchical} to improve the performance of fashion compatibility learning.
Unlike simply determining whether an outfit is compatible or not, determining why an outfit is compatible or not focuses more on the explanation of compatibility.
Wang \textit{et al.} \cite{wang2019diagnosis} first proposed a diagnosis module using a first-order Taylor expansion to find the salient reason for compatibility.
Subsequently, Yang \textit{et al.}\cite{yang2021attribute} employed an attention network to find the reason by considering attributes.
Apart from tackling the above two key issues in outfit recommendation, creating a compatible outfit in a generative way has also drawn attention from researchers.
In particular, Liu \textit{et al.}\cite{liu2019toward} first proposed an Attribute-GAN model to synthesize complementary fashion items in both the upper and lower clothing domains, i.e., synthesizing complementary upper clothing based on given lower clothing or vice versa.
Afterwards, Liu \textit{et al.}\cite{liu2019collocating} further improved their GAN model by developing a multi-discriminator structure.
Yu \textit{et al.} \cite{yu2019personalized} designed another GAN model by considering personality.
Hsiao \textit{et al.} \cite{hsiao2019fashion++} enhanced the fashionability of an outfit in a minimal edit for compatibility improvement strategy.

\textbf{Image-to-Image Translation}.
Image-to-image translation aims at learning a conditional distribution from one given image domain to another image domain.
Image-to-image translation models can be divided into supervised \cite{pix2pix2017, wang2018pix2pixHD} and unsupervised methods\cite{huang2018munit,CycleGAN2017,park2020cut}.
The aim of supervised methods is to translate images from one domain to another with paired images.
Isola \textit{et al.} \cite{pix2pix2017} proposed a Pix2Pix model, which uses GANs and the L1 (or L2) loss, to accomplish supervised image-to-image translation with ground truths.
Wang \textit{et al.} \cite{wang2018pix2pixHD} further improved Pix2Pix with a coarse-to-fine strategy.
On the other hand, unsupervised methods accomplish image-to-image translation without using paired images.
Zhu \textit{et al.} \cite{CycleGAN2017} introduced a cycle reconstruction loss with two-directional generators to learn the distributions of two different domains and preserve their original domain information.
Then Huang \textit{at al.} \cite{huang2018munit} disentangled the images into content code and style code in latent space so as to improve the performance of unsupervised methods.
Recently, Park \textit{et al.} \cite{park2020cut} have used self-supervised learning to further improve unsupervised methods. Furthermore, Li \textit{et al.}\cite{li2021image} utilized labels under a hierarchical tree structure to improve the image-to-image translation performance, especially in controllable attribute editing. Other modalities such as sketches\cite{li2020staged} and textual descriptions \cite{yang2021multi} have also been explored for synthesizing photo-realistic images based on their conditional inputs.

\textbf{Features of Our Work}.
This research focuses on synthesizing complementary fashion items based on arbitrary given fashion items to compose a compatible outfit, called ``outfit generation.''
Outfit generation can be used to generate compatible fashion items and assemble them into outfit recommendations.
Unlike previous work on synthesizing compatible fashion items \cite{liu2019toward, liu2019collocating, yu2019personalized} solely in the upper and lower clothing domains, our outfit generation framework concentrates more on generating compatible fashion items based on given arbitrary fashion items from multiple clothing domains.
On the other hand, as mentioned, our task can be seen as one special kind of image-to-image translation task.
Compared with the general image-to-image translation task \cite{pix2pix2017, wang2018pix2pixHD, huang2018munit,CycleGAN2017,park2020cut}, the number of input images in our framework can be a variable value, and the input and the desired output of the fashion images have no explicit spatial alignment.
To address these issues, we have developed a new outfit generation framework, COutfitGAN, which includes a pyramid style extractor for effectively encoding arbitrary given fashion items, an outfit generator for adaptively fusing the silhouette and style information of the given fashion items, and a collocation discriminator for assessing the outfit compatibility of these items with the given ones in a supervised manner.

\section{Method}
\label{method}
This section introduces the outfit generation framework, which uses a pyramid style extractor, an outfit generator, a real/fake discriminator, and a collocation discriminator.
First, the problem will be given a formal description.
Then we will give the motivation and an overview of our developed COutfitGAN.
The pyramid style extractor, outfit generator, real/fake discriminator, collocation discriminator, training losses, and adversarial training algorithms will be described in sequence.
\subsection{Problem Formulation}
\label{pro_forl}
In general, previous outfit recommendation methods \cite{liu2019toward,liu2019collocating,yu2019personalized} for synthesizing new compatible fashion items mainly focus on the upper and lower clothing domains.
They formulate their problem as an image-to-image translation task that takes a fashion item's image as input and produces its compatible item's image.
In contrast, we extend upper--lower clothing generation to generating outfits based on arbitrary fashion items by synthesizing composable and compatible fashion items.
We formulate this outfit generation problem as follows.
For an outfit $\mathcal{O}$, suppose that it includes $N$ fashion items, i.e., $\{\mathcal{O}_1, \cdots, \mathcal{O}_i, \cdots, \mathcal{O}_N\}$.
Each fashion item $\mathcal{O}_i$ belongs to a unique category in $\mathcal{O}$, and these categories are distinct from each other.
This means that $N$ fashion items are collected from $N$ categories to compose a complete outfit.
For a given set with $m$ fashion items $\{\mathcal{O}_{g_1},\cdots,\mathcal{O}_{g_m}\}$, our outfit generation framework needs to synthesize a complementary set with $n$ fashion items $\{\widetilde{\mathcal{O}_{s_1}},\cdots,\widetilde{\mathcal{O}_{s_n}}\}$ where $m+n=N$ and $\{g_1,\cdots,g_m,s_1,\cdots,s_n\}=\{i\}^N_{i=1}$.
Each fashion item $\mathcal{O}_i$, where $1\leq i \leq N$, is associated with a silhouette mask $Sl_i$.
Here, the silhouette masks may be given by a user, or selected by the user from a candidate dataset containing various silhouettes of fashion items.
Our goal is to learn a generative model that aims at synthesizing complementary fashion items based on arbitrary given fashion items to compose a compatible outfit.
\subsection{Motivation of Our Method}
The aim of this research is to synthesize a set of complementary fashion items based on a set of given fashion items, while the synthesized fashion items should be photo-realistic and compatible with the given ones. For clarity, we describe the motivation of our method from the following perspectives.

Firstly, style-based GANs (StyleGAN\cite{karras2019style} and StyleGAN2\cite{karras2020analyzing}) have achieved a great success in unconditional image generation in recent years. In essence, their generation module includes two main parts, i.e., a mapping network that maps Gaussian noise to a disentangled style embedding and a synthesis network that takes the style embedding to manipulate the feature map repeatedly $18$ times to synthesize $1024\times 1024$ images in a progressive way. The disentangled style embedding lies in a space of $\mathbb{R}^{1\times512}$ which is called $\mathcal{W}$ space. 
Then many StyleGAN-based network structures\cite{abdal2019image2stylegan,richardson2021encoding,tov2021designing} were applied to conditional image generation and have shown better performance than the encoder--decoder structure in image-to-image translation. These methods extended the original $\mathcal{W}$ space into a $\mathcal{W}+$ space, where $\mathcal{W}+\subseteq\mathbb{R}^{18\times512}$ is defined by the concatenation of 18 different 512-dimensional style embeddings. They used $18$ different style embeddings to manipulate the feature map in a progressive way. To take advantage of the power of StyleGAN, we adapt its generation module to our research problem from two perspectives. On the one hand, previous StyleGAN-based methods\cite{abdal2019image2stylegan,richardson2021encoding,tov2021designing} first need to map images to the $\mathcal{W}+$ space. These style embeddings are then edited and fed into the synthesis network to synthesize the targeted images. In our problem setting, we cannot obtain the style embedding from their original images directly. The style embeddings of complementary fashion items can only be generated based on the given ones. In a previous work, Han \textit{et al.} \cite{han2017learning} proposed that the fashion items in an outfit can be seen as a sequence from a human observation perspective and a bi-directional long short-term memory (Bi-LSTM) was used to predict the complementary fashion items conditioned on the given ones. In line with these works, we design a new pyramid style extractor that includes a CNN backbone for visual feature extraction of given fashion items and a Bi-LSTM module to generate the style embeddings of the complementary fashion items. Following the work \cite{richardson2021encoding}, we employ a three-stage CNN backbone to extract the visual embeddings from low-, middle- and high-level perspectives. Different from their CNN backbone, here we only use the CNN backbone to extract the visual embeddings of the given fashion items. The style embeddings of the complementary fashion items are generated by the Bi-LSTM module which takes the visual embeddings and their categorical embeddings of given fashion items as input. On the other hand, these aforementioned StyleGAN-based methods can synthesize images only based on style embeddings. Another study \cite{10.1145/3306346.3323023} reported that only utilizing style embeddings cannot have good spatial control for synthesizing images under the StyleGAN framework. Meanwhile, it is difficult to model the spatial mapping from given fashion items to complementary ones due to the uncertainty of image-to-image translation in our problem setting. Hence, we design an outfit generator including some silhouette and style (S-S) fusion blocks to reduce the uncertainty of the spatial mapping as well as to achieve better control.
Secondly, the synthetic images of fashion items are expected to be photo-realistic. To improve the visual authenticity of the synthesized images, we employ a UNet-based discriminator \cite{schonfeld2020u}, which is capable of supervising the synthetic results from both global and local viewpoints.
Finally, our objective is to synthesize a set of fashion items that are compatible with the given ones. To achieve this, we propose a new collocation discriminator that regards the whole outfit as a sentence and each fashion item in this outfit as a word in the sentence inspired by the field of natural language processing (NLP). Specifically, all fashion items are first mapped into a common compatibility space that regards each item as an embedding. Then a contrastive learning strategy is used to construct a collocation discriminator by exploiting the in-between relationships of the fashion items with their embeddings.

\subsection{An Overview of COutfitGAN}
\begin{figure}[t]
	\centering
	\includegraphics[width=0.49\textwidth, height=0.3748352\textwidth]{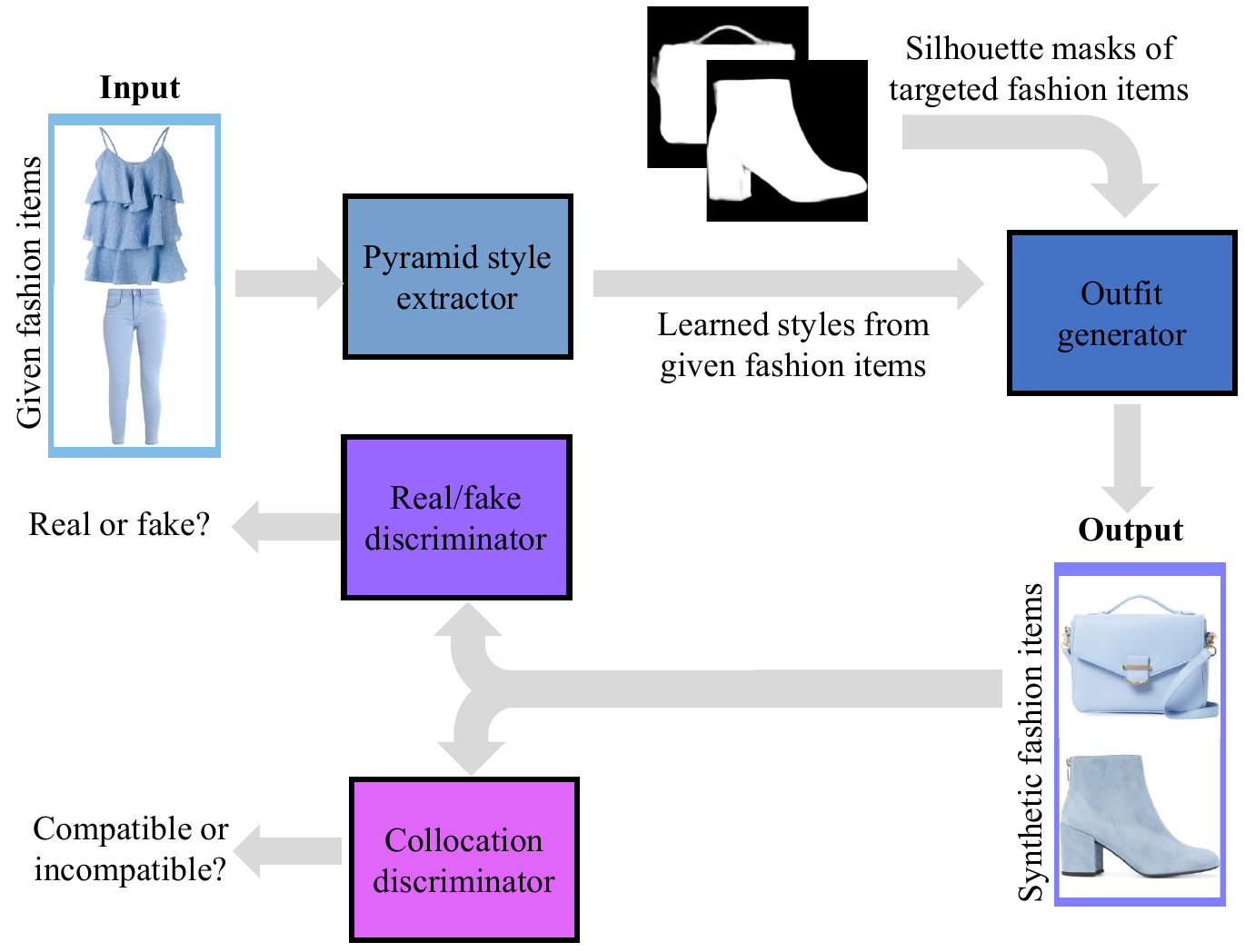}
	\caption{Overview of our COutfitGAN framework.}
	\label{overview}
	\vspace{-0.5cm}
\end{figure}
Based on the above problem formulation, we have designed a new generative framework, named COutfitGAN, to fulfill the task of outfit generation.
As shown in Fig.~\ref{overview}, the overall framework of COutfitGAN includes four parts: a pyramid style extractor, an outfit generator, a real/fake discriminator, and a collocation discriminator.
More specifically, the pyramid style extractor is responsible for extracting the styles of the given fashion items in such a way that the generation of complementary fashion items can learn from these styles and preserve them during  synthesis.
The outfit generator progressively fuses silhouette masks and these learned styles of the targeted fashion items to synthesize the complementary items.
The real/fake discriminator and collocation discriminator supervise the inference of synthesized images in terms of visual authenticity and outfit compatibility, respectively.
In the following, the details of the implementation of each key module will be described in turn.

\subsection{The Pyramid Style Extractor}
\begin{figure*}[t]
	\centering
	\includegraphics[width=0.95\textwidth, height=0.32787611\textwidth]{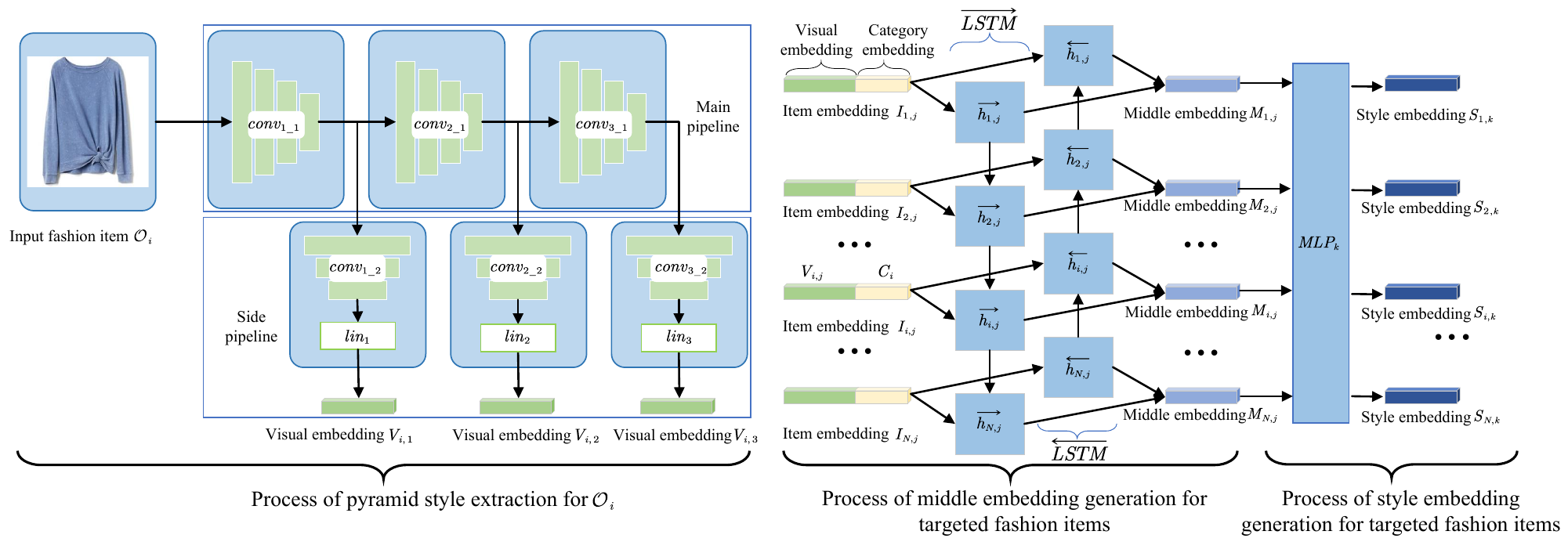}
	\caption{The architecture of the pyramid style extractor.}
	\label{pyramid_style_extractor}
\end{figure*}
To synthesize a set of complementary items to make up an outfit, our framework needs to learn the style representations of complementary items based on the given fashion items, by considering the hidden style interaction between fashion items.
An early study \cite{han2017learning} reported that fashion items in an outfit can be seen as a sequence from a human's scanning, that is, humans always observe other people from up to down or in the opposite direction.
On the other hand, Kim \textit{et al.}\cite{kim2021self} proved that textures play an important role in fashion compatibility learning, and previous works \cite{heeger1995pyramid,gatys2015texture} indicated that pyramid features turn out to be essential in texture analysis or synthesis.
Inspired by these prior studies, we have designed a pyramid style extractor that is used in outfit generation to characterize fashion style representations for complementary fashion items.
These fashion style representations are also suitable for plugging into our outfit generator, which is designed under the StyleGAN2 \cite{karras2020analyzing} architecture (see details in Section~\ref{outfit_generator_}) for synthesizing fashion items.
In particular, as shown in Fig.~\ref{pyramid_style_extractor}, our pyramid style extractor includes three modules: a CNN backbone for feature extraction, a Bi-LSTM structure \cite{bi_lstm} for generating the features of the targeted fashion items, and a multi-layer perceptron (MLP) network for style mapping.
Considering that pyramid features can be easily transplanted into StyleGAN2 due to its progressive input structure, we designed a pyramid style extractor with a multi-scale strategy.
Our pyramid style extractor plays a similar role as the mapping module in StyleGAN2.
However, unlike the mapping module, in which all the information is provided for the existing styles in advance, we need to first generate the complementary styles based on the given fashion items.
For the first stage of feature extraction, we use a three-stage CNN to extract pyramid features at three scales.
As shown in Fig.~\ref{pyramid_style_extractor}, the CNN feature extraction module includes a main pipeline and a side pipeline in three scales.
We denote the CNN modules in the main pipeline in sequence by $ conv_{1\_1}$, $conv_{2\_1}$, and $conv_{3\_1}$.
The corresponding CNN modules in the side pipeline are denoted by $ conv_{1\_2}$, $conv_{2\_2}$ and $conv_{3\_2}$, and the linear projection modules in the side pipeline are denoted by $lin_1$, $lin_2$ and $lin_3$.
Let $f_{i,j}$ ($j\in\{1,2,3\}$ in this research) denote the $i$-th fashion item's $j$-th scale feature extracted from the CNN module in the main pipeline.
Formally, these operations are described as follows:
\begin{equation}
\begin{cases}
		f_{i,0}=\mathcal{O}_i,& \\
	f_{i,j} = conv_{j\_1}(f_{i,j-1}),              & \text{if}\ j\in \{1,2,3\}.
\end{cases}
\end{equation}

Then we use a side CNN and a linear projection layer in the side pipeline to map these features into an embedding space.
These operations are described in the following form:
\begin{equation}
	V_{i,j} = lin_{j}[conv_{j\_2}(f_{i,j})],
\end{equation}
where $conv_{j\_2}$ and $lin_{j}$ denote the above CNN and linear projection layer in the side pipeline, respectively.
In particular, when $\mathcal{O}_i$ is not a given fashion item, $V_{i,j}$ becomes a zero vector.
In line with \cite{orel2020lifespan}, these visual embeddings $V_{i,j}$ are then concatenated with a ($D_{cat}\times N$)-dimensional category embedding (here, $D_{cat}$ is set to $50$, which is the same setting as in \cite{orel2020lifespan}) to form the $i$-th fashion item in the $j$-th scale embedding $I_{i,j}$.
After feature extraction for the given fashion items, we need to generate the style embedding for the targeted fashion items.
Then the item embeddings $\{I_{i,j}\}_{i=1}^N$ are fed into a Bi-LSTM module to generate the middle embeddings $\{M_{i,j}\}_{i=1}^{N}$ of the targeted fashion items with this sequential model.
Taking the forward direction of the Bi-LSTM as an example, these operations are described as follows:

\begin{equation}
	I_{i,j}=V_{i,j}\oplus C_{i},
\end{equation}
where $\oplus$ denotes a concatenation operation and $C_{i}$ is the aforementioned ($D_{cat}\times N$)-dimensional category embedding.
Then the item features are fed into the Bi-LSTM module.
Taking the forward process as an example, it can be formulated as follows:
\begin{equation}
	\overrightarrow{{h}_{i,j}} = \overrightarrow{LSTM}(I_{1,j}, I_{2,j},\cdots,I_{i,j}),
\end{equation}
where $\overrightarrow{LSTM}$ and $\overrightarrow{{h}_{i,j}}$ denote the forward LSTM and its $i$-th hidden feature, respectively.

In line with \cite{karras2019style,karras2020analyzing}, we add three MLPs to  formulate a better style embedding space for low-, middle-, and high-level perspectives.
Due to the progressivity of style control, it has become common practice \cite{richardson2021encoding} to synthesize images with an extended latent space, $\mathcal{W}+$, denoted by the concatenation of 18 different 512-dimensional vectors for synthesizing images in $\mathbb{R}^{1024\times1024\times3}$.
Here, each vector is responsible for the corresponding input layer of StyleGAN2\cite{karras2020analyzing}.
To synthesize an image lying in $\mathbb{R}^{256\times256\times3}$, we need $14\times 512~\mathcal{W}+$ to adaptively adjust the styles of synthesized fashion items.
To achieve the above goal, each MLP includes $N_{mlp}$ layers.
Each layer has $N_{neun}$  neurons except for the last layer.
For the last layer, these MLPs have five, four, and five heads, where each head has $512$ neurons. Let $MLP_k$ denote the $k$-th ($k\in \{1,\cdots,14\}$) head and its related ($N_{mlp}-1$) layers of the MLPs as mentioned above.
Each middle embedding $M_{i,j}$ is fed into an MLP for a style embedding $S_{i,k}$.
These middle feature embeddings are fed into three different MLPs to generate the first $5\times512 ~ \mathcal{W}+$, the middle $4\times512 ~ \mathcal{W}+$, and the last $5\times512 ~ \mathcal{W}+$ of the total $14 \times 512 ~\mathcal{W}+$.
Each middle embedding $M_{i,j}$ is obtained as follows:

\begin{equation}
	M_{i,j} = \overleftarrow{{h}_{i,j}} \oplus \overrightarrow{{h}_{i,j}},
\end{equation}
where $\overleftarrow{{h}_{i,j}}$ and $\overrightarrow{{h}_{i,j}}$ denote the $i$-th hidden features of backward and forward LSTMs, respectively.

Then the middle embedding $M_{i,j}$ ($j=\{1,2,3\}$) is fed into the MLPs for generating a total of $14\times512 ~\mathcal{W}+$, where all three MLPs include $14$ heads to generate a total of $14\times512 ~\mathcal{W}+$.
This operation is formulated as follows:
\begin{equation}
	S_{i,1},\cdots, S_{i,k},\cdots, S_{i,14} = MLP(M_{i,1}, M_{i,2}, M_{i,3}),
\end{equation}
where $S_{i,k}$ denotes the final $k$-th style embedding for the $i$-th fashion item in the $k$-th layer.
These pyramid style embeddings are fed into the outfit generator to synthesize complementary fashion items with silhouette mask information.
The first-, second-, and third-scale features are mapped into the first five layers, the middle four layers, and the last five layers, for a total of $14 \times 512 ~ \mathcal{W}+$ vectors.
Then these latent codes are progressively fed into the generation module to produce the final synthetic images.
 
\subsection{The Outfit Generator}
\begin{figure}[t]
	\centering
	\includegraphics[width=0.49\textwidth, height=0.42348416\textwidth]{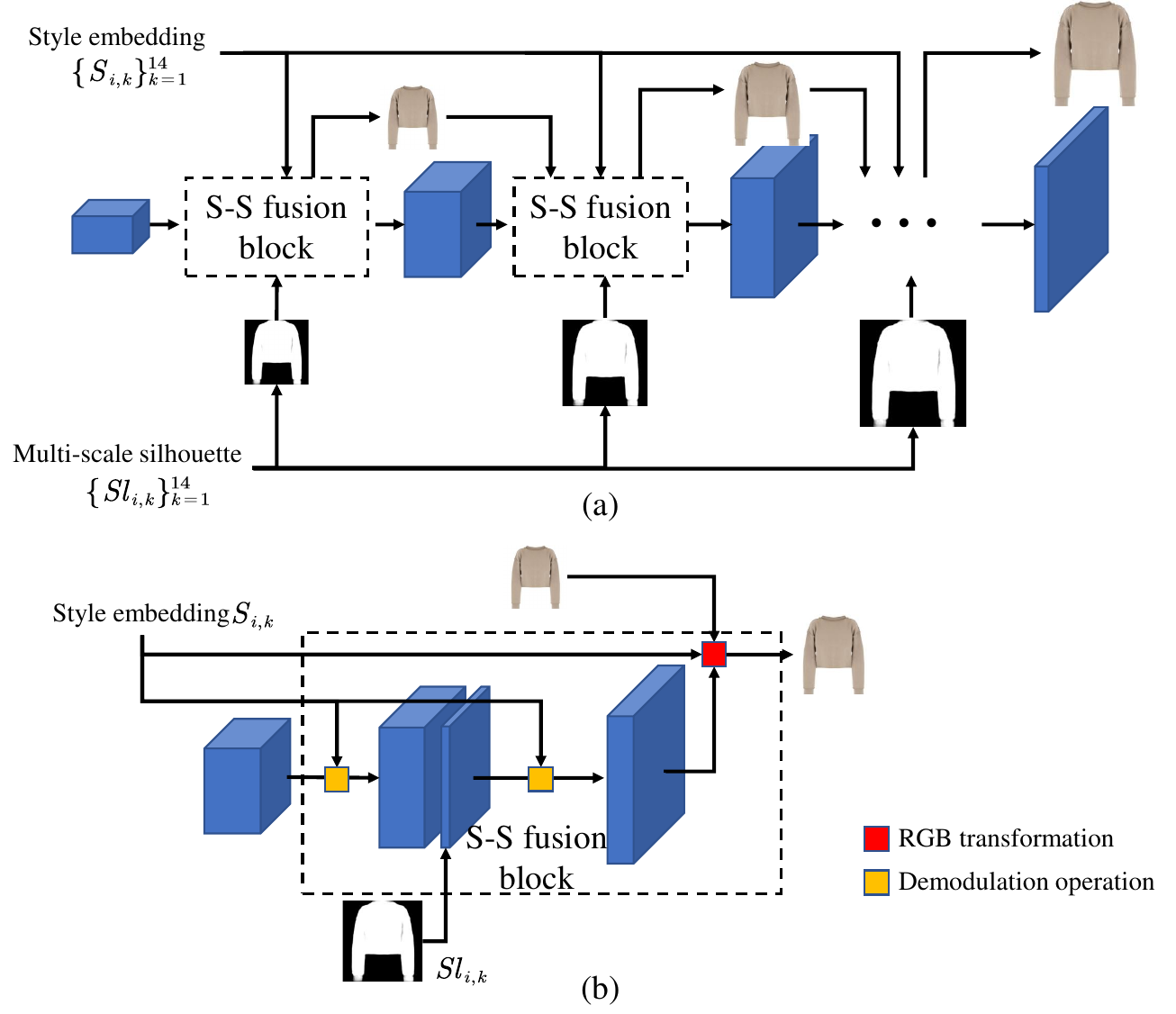}
	\caption{The architecture of the outfit generator: (a) structure overview, and (b) silhouette and style (S--S) fusion block.}
	\label{outfit_generator}
	\vspace{-0.5cm}
\end{figure}
\label{outfit_generator_}
Inspired by the great success of progressive generation techniques \cite{karras2017progressive,karras2019style,karras2020analyzing}, we adapt the generator from StyleGAN2\cite{karras2020analyzing},
which contains a mapping module and a progressive generation module.
Corresponding to the generation module from StyleGAN2, we re-designed the outfit generator in our framework: the reference silhouette masks and styles from the pyramid style extractor are fused, as shown in Fig.~\ref{outfit_generator} (a).
Our outfit generator includes a number of silhouette and style (S--S) fusion blocks, which fuse all silhouette masks and styles to progressively synthesize the corresponding fashion items.
For each fusion block, as shown in Fig.~\ref{outfit_generator} (b), every silhouette mask is downsampled and concatenated into the current feature map.
Each style embedding $S_{i,k}$ is first fed into an affine transformation, and then the features processed by the former CNN are demodulated. Weight demodulation is an operation adopted by StyleGAN2 to demodulate the weights of convolutional layers, helping to manipulate the feature map with the style embedding.
Formally, these operations are described as follows:

\begin{equation}
	S_{i,k}' = \mathcal{A}(S_{i,k}),
\end{equation}
where $\mathcal{A}$ is the affine transformation mentioned above.
Subsequently, each transformed style embedding controls the style of the feature map produced by the convolutional layer by modulating the parameter weights $w$ of the convolutional layer. Each $w$ is a three-dimensional tensor, including the channel, height and width dimensions. Weight demodulation is applied to the channel dimension to demodulate $w$ as follows:
\begin{equation}
	w'_{j_1,j_2,j_3} = S'_{i,k}@{j_1}\cdot w_{j_1,j_2,j_3},
\end{equation}
where $S'_{i,k}@{j_1}$ denotes the value of $S'_{i,k}$ at the $j_1$-th position, $w$ are the original weights, and $w'$ are the modulated weights.
Then these weights are also scaled through an L2 norm to achieve normalization in the same manner as in \cite{karras2020analyzing}.
\subsection{The Real/Fake Discriminator}

\label{real_or_fake_dis}
Unlike natural images, fashion item images have many white blanks on the border of the canvas, which makes these images usually not continuous in intensity.
We need to supervise the image generation from the pixel level for better synthesis.
Our real/fake discriminator employs a UNet-based discriminator \cite{schonfeld2020u} to supervise the generation of the synthesized images in terms of their visual authenticity, as shown in Fig.~\ref{discriminator}.
This architecture serves as the COutfitGAN discriminator network, which returns both a scalar and a pixel-wise real/fake classification result.
According to our observations, this discriminator can greatly enhance the local and global visual appearance of the synthesized fashion images.
To make this paper self-contained, we describe the loss for the UNet-based discriminator as follows:
\begin{figure}[t]
	\centering
	\includegraphics[width=0.49\textwidth, height=0.30053333\textwidth]{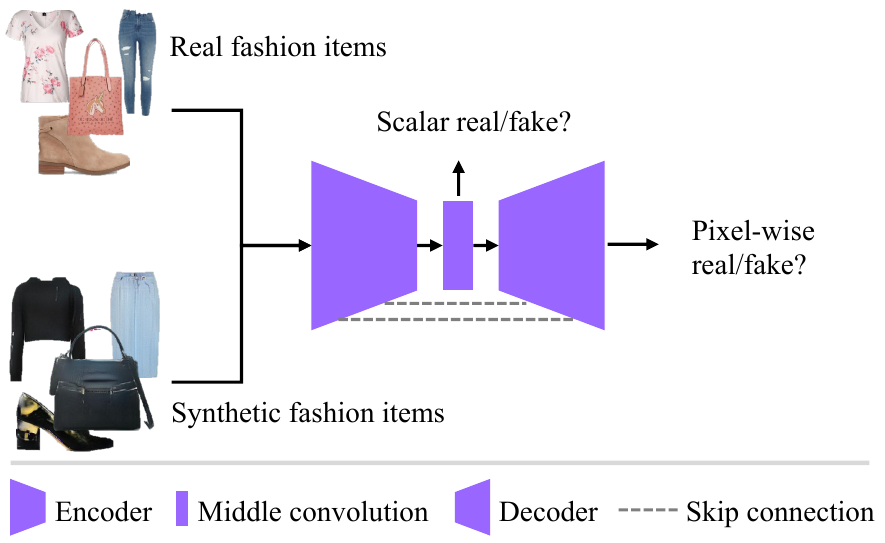}
	\caption{The UNet-based real/fake discriminator.}
	\label{discriminator}
	\vspace{-0.5cm}
\end{figure}
\begin{equation}
	\mathcal{L}_{dis}=\mathcal{L}_{D_{enc}}+\mathcal{L}_{D_{dec}},
\end{equation}
where $\mathcal{L}_{dis}$ denotes the loss function for training our real/fake discriminator, $\mathcal{L}_{D_{enc}}$ is the scalar real/fake loss, and $\mathcal{L}_{D_{dec}}$ represents the pixel-wise loss.
In addition, $D_{enc}(\cdot)$ and $D_{dec}(\cdot)$ denote the functions after the encoder and decoder, respectively.
For clarity, we describe the details of these two loss functions as follows:
\begin{equation}
	\begin{aligned}
		\mathcal{L}_{D_{enc}} = &-\mathbb{E}_{\mathcal{O}_i} [\log [D_{enc}(\mathcal{O}_i)]]\\
		&-\mathbb{E}_{\widetilde{\mathcal{O}}_i}[ \log (1-[D_{enc}(\widetilde{\mathcal{O}}_i)]],
	\end{aligned}
\end{equation}
\begin{equation}
	\begin{aligned}
		\mathcal{L}_{D_{dec}} = &-\mathbb{E}_{\mathcal{O}_i}\sum_{h,w}[\log [D_{dec}(\mathcal{O}_i)]_{h,w}]\\
		&-\mathbb{E}_{\widetilde{\mathcal{O}}_i}[\sum_{h,w}\log (1-[D_{dec}(\widetilde{\mathcal{O}}_i))]_{h,w})],
	\end{aligned}
\end{equation}
where $\mathcal{O}_i$ denotes the $i$-th real fashion item, $\widetilde{\mathcal{O}}_i$ denotes the $i$-th synthesized item, and $h$ and $w$ denote the height and width indexes for the feature map from $D_{dec}$.

For the generation module, the GAN's loss function from the UNet-based discriminator is formulated as follows:
\begin{equation}
	\label{l_gan}
	\mathcal{L}_{gan}=-\mathbb{E}_{\widetilde{O}_i}[\log D_{enc}(\widetilde{O}_i)+\sum_{h,w}\log [D_{dec}(\widetilde{O}_i)]_{h,w}].
\end{equation}
With such a loss function, the visual authenticity of the synthesized images can be improved from global and local perspectives during adversarial training.
\subsection{The Collocation Discriminator}

\label{coll_dis}
\begin{figure}[t]
	\centering
	\includegraphics[width=0.49\textwidth, height=0.24704167\textwidth]{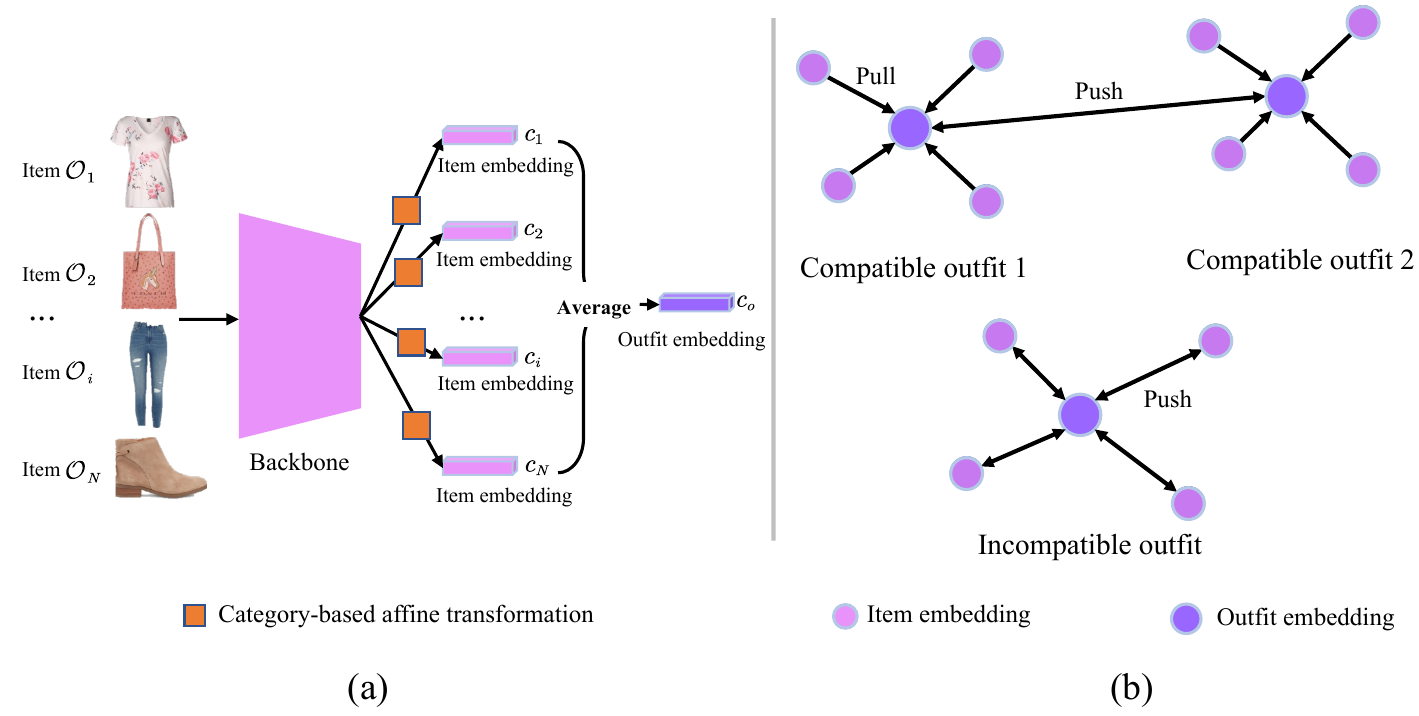}
	\caption{The mechanism of the collocation discriminator: (a) fashion item and outfit embedding in the compatibility space through a CNN backbone and a category-based affine transformation and (b) the collocation discriminator with contrastive learning.}
	\label{collocation_discriminator}
	\vspace{-0.5cm}
\end{figure}

In the field of NLP, a sentence vector can be obtained by an averaging operation on every word embedding\cite{le2014distributed}.
Inspired by this, we believe that for an outfit, the outfit embedding for representing a global style can also be achieved by an averaging operation over all of the fashion item embeddings, each of which can be regarded as a local style in an outfit.
Firstly, as shown in Fig.~\ref{collocation_discriminator} (a), we map each fashion item into the same fashion compatibility space through a CNN backbone and a category-based affine transformation.
For each outfit, let the average of the style embeddings of all the fashion items represent the whole outfit's style.
The category-based affine transformation can be represented by an $N\times D$ matrix.
Formally, the definition of this operation is
\begin{equation}
	c_i = C'_i \times M_{c} \times F_i,
\end{equation}
where $C'_i$ is a category vector lying in ($1\times N$)-dimensional space with one-hot encoding; $M_{c}$ is the matrix of a learnable category-based affine transformation; $F_i$ is the $i$-th fashion item's embedding extracted by the CNN backbone; and $c_i$, which denotes the $i$-th fashion item's style, is the $i$-th fashion item's embedding in the common compatibility space. The outfit embedding $c_o$, which denotes the global style of an outfit, can be obtained by an averaging operation over all fashion item embeddings in the common compatibility space.
Then, we employ contrastive learning to supervise the compatibility of the synthesized fashion items.
Intuitively, as shown in Fig.~\ref{collocation_discriminator} (b), we expect that different outfits should have large in-between distances in the global outfit style embedding, and the embedding of each fashion item in a compatible outfit should have a small distance from the global outfit style embedding, and vice versa.
To achieve this, we use the following loss function for training our collocation discriminator:
\begin{equation}
	\label{l_coll_dis}
	\mathcal{L}_{coll_{dis}} = \mathbb{E}\sum_{i=1}^{N}||c_i^p-c_o^p||_2^2-\mathbb{E}\sum_{i=1}^{N}||c_i^n-c_o^n||_2^2-\mathbb{E}||c_o^{p1}-c_o^{p2}||_2^2,
\end{equation}
where $c_i^p$ denotes the $i$-th fashion item from a compatible outfit style embedding in the compatibility space and $c_o^p$ is computed by $\frac{1}{N}\sum_{i=1}^{N}c_i^p$; similarly, $c_i^n$ denotes the $i$-th fashion item from an incompatible outfit style embedding in the compatibility space, and $c_o^n$, computed by  $\frac{1}{N}\sum_{i=1}^{N}c_i^n$, gives the outfit style embedding in the compatibility space; in addition, $c_o^{p1}$ and $c_o^{p2}$ denote different outfit style embeddings for two different compatible outfits.
The last term of the loss functions is used to make the collocation discriminator supervise the generator to synthesize diverse styles during training.

Correspondingly, we hope that the synthesized fashion items should remain compatible in a composed outfit.
Thus, the objective function of the collocation discriminator becomes:

\begin{equation}
	\mathcal{L}_{coll} = \mathbb{E}\sum_{i=1}^{N}||{c'}_i^p-{c'}_o^p||_2^2-\mathbb{E}\sum_{i=1}^{N}||{c'}_i^n-{c'}_o^n||_2^2,
\end{equation}
where ${c'}_i^p$ denotes the synthesized $i$-th fashion item from a compatible outfit style embedding in the compatibility space if this fashion item was not given, otherwise, it denotes the $i$-th given fashion item style embedding in the compatibility space.
\subsection{Training losses}
Our training loss functions can be divided into three parts separately: one for the pyramid style extractor and outfit generator, one for the real/fake discriminator, and one for the collocation discriminator.
Those for the real/fake and collocation discriminators are described in Sections \ref{real_or_fake_dis} and \ref{coll_dis}.
Here we only describe our training loss functions for the pyramid style extractor and outfit generator.
These two modules are trained together with the same loss functions:

\begin{equation}
	\label{total_loss}
	\mathcal{L}_g = \mathcal{L}_{gan} + \lambda_1 \mathcal{L}_1 + \lambda_2 \mathcal{L}_{vgg} + \lambda_3 \mathcal{L}_{coll},
\end{equation}
where $\lambda_1$, $\lambda_2$, and $\lambda_3$ are predefined hyper-parameters of tradeoffs of all losses. Here, $\mathcal{L}_g$ is responsible for supervising the COutfitGAN to synthesize photo-realistic images; $\mathcal{L}_1$ and $\mathcal{L}_{vgg}$ are responsible for supervising the synthesized results of COutfitGAN that have a small distance with the ground truths at the pixel and feature levels, respectively; and our proposed $\mathcal{L}_{coll}$ is responsible for supervising the synthesized fashion items that can maintain compatibility with the existing items.

\textbf{L1 Loss}: To minimize the difference between the target outfits and the synthesized ones, we use the L1 loss ($\mathcal{L}_1$) to capture the overall structure of the images from the target domain.
The L1 loss calculates the absolute distance between the synthesized images and the target images \cite{pix2pix2017}.
It is defined as follows:

\begin{equation}
	\mathcal{L}_1 = \mathbb{E}[\frac{1}{N}\sum_{i=1}^{N}||\mathcal{O}_{i}-\widetilde{\mathcal{O}_{i}}||_1],
\end{equation}
where $N$ is the number of fashion items in an outfit and $\widetilde{\mathcal{O}_{i}}$ is the synthesized $i$-th fashion item.

\textbf{VGG Loss}: Unlike the L1 loss, the VGG loss ($\mathcal{L}_{vgg}$)\cite{ledig2017photo} is introduced to ensure that the synthesized images are close to the target ones in high-level feature space.
It also measures the perceptual difference between the images in terms of their content and style.
Here, we adopt the VGG loss to ensure that our COutfitGAN produces images that are similar to the ground truths.
We compute the VGG loss in the 
\textit{relu1\_2}, \textit{relu2\_2}, \textit{relu3\_3} and \textit{relu4\_3} layers of the VGG-16 \cite{simonyan2014very} in the following form:
\begin{equation}
	\mathcal{L}_{vgg} = \mathbb{E}[\frac{1}{N}\sum_{i=1}^{N}\sum_l||\phi_l(\mathcal{O}_{i})-\phi_l (\widetilde{\mathcal{O}_{i}})||_1],
\end{equation}
where $l\in \{relu1\_2,relu2\_2,relu3\_3,relu4\_3\}$ is the aforementioned layer of VGG-16, and $\phi _l$ represents the function of layer $l$.
\subsection{The Adversarial Training Process}

\begin{algorithm}[t]
	\SetKwFunction{FMain}{SynOutfit}
	\SetKwProg{Fn}{Function}{:}{}
	\small
	\caption{Adversarial training algorithm for COutfitGAN}
	\label{algo}
	\SetAlgoLined
	\KwIn{Compatible outfit $\mathcal{O}$ which includes $N$ fashion items, let $\mathcal{O}_i$ denote the $i$-th fashion item in $\mathcal{O}$.
		Each fashion item $\mathcal{O}_i$ is associated with a silhouette mask $Sl_i$.}
	\KwOut{Pyramid style extractor $E$ and generator $G$ of COutfitGAN}
	Initialize the parameters ($\theta_{E}$, $\theta_{G}$, $\theta_D$, and $\theta_{D_{coll}}$) of $E$, $G$, $D$, and $D_{coll}$, respectively\;
	\For{$iter\leftarrow 1$ \KwTo $N_{iter}$}{
		sample a batch of $\mathcal{O}=[\mathcal{O}_{1},\cdots,\mathcal{O}_{N}]$ and corresponding silhouette masks $[Sl_{1},\cdots,Sl_{N}]$ from the training set\;
		sample a batch of masks for fashion items $[Mask_1,\cdots,Mask_N]$ to construct given fashion items and targeted fashion items, where $Mask_i\in\{0,1\}$, and no less than $1$ fashion item(s) are given\;
		fix $\theta_E$, $\theta_G$, and $\theta_{D_{coll}}$, set $\theta_D$ learnable\;
		$[\widetilde{\mathcal{O}}_1,\cdots,\widetilde{\mathcal{O}}_N]$ $\leftarrow$ \texttt{SynOutfit(}$[\mathcal{O}_{1},\cdots,\mathcal{O}_{N}]$, $[Sl_1, \cdots, Sl_N]$, $[Mask_1,\cdots,Mask_N]$\texttt{)}\;
		$\mathcal{L}_{gan}\leftarrow -\mathbb{E}_{\widetilde{O}_i}[\log D_{enc}(\widetilde{O}_i)+\sum_{h,w}\log [D_{dec}(\widetilde{O}_i)]_{h,w}]$; // See Eq.
		(\ref{l_gan}) \\
		update $\theta_D$ with \\
		\quad $\theta_D\leftarrow \theta_D - \eta \bigtriangledown_{\theta_D}$\;
		fix $\theta_E$, $\theta_G$, and $\theta_{D}$, set $\theta_{D_{coll}}$ learnable\;
		$\mathcal{L}_{coll_{dis}} = \mathbb{E}\sum_{i=1}^{N}||c_i^p-c_o^p||_2^2-\mathbb{E}\sum_{i=1}^{N}||c_i^n-c_o^n||_2^2-\mathbb{E}||c_o^{p1}-c_o^{p2}||_2^2$; // See Eq.~(\ref{l_coll_dis})\\
		update $\theta_{D_{coll}}$ with\\
		\quad $\theta_{D_{coll}}\leftarrow \theta_{D_{coll}} - \eta \bigtriangledown_{\theta_{D_{coll}}}$\;
		fix $\theta_{D}$ and $\theta_{D_{coll}}$, set $\theta_E$ and $\theta_G$ learnable \;
		$\mathcal{L}_g = \mathcal{L}_{gan} + \lambda_1 \mathcal{L}_1 + \lambda_2 \mathcal{L}_{vgg} + \lambda_3 \mathcal{L}_{coll}$; // See Eq.~(\ref{total_loss})\\
		update $\theta_{E}$ and $\theta_{G}$ with\\
		\quad $\theta_{E}\leftarrow \theta_{D_{E}} - \eta \bigtriangledown_{\theta_{E}}$\;
		\quad $\theta_{G}\leftarrow \theta_{G} - \eta \bigtriangledown_{\theta_{G}}$\; 
	}
	\Fn{\FMain{$[\mathcal{O}_{1},\cdots,\mathcal{O}_{N}]$, $[Sl_1, \cdots, Sl_N]$, $[Mask_1,\cdots,Mask_N]$}}{
		$S_{i,k} = E(\mathcal{O}_1\cdot Mask_1, \cdots, \mathcal{O}_N\cdot Mask_N)$, where $i\in \{1,\cdots,N\}$ and $k\in\{1,\cdots,14\}$\;
		$\widetilde{\mathcal{O}}_i=G([S_{i,1},\cdots,S_{i,14}], Sl_i)$, where $i\in \{1,\cdots,N\}$\;
		$\widetilde{\mathcal{O}}_i\leftarrow \widetilde{\mathcal{O}}_i \cdot (1-Mask_i) + \mathcal{O}_i \cdot Mask_i$\ , where $i\in \{1,\cdots,N\}$\;
		\textbf{return} $[\widetilde{\mathcal{O}}_1,\cdots,\widetilde{\mathcal{O}}_N]$\; 
	}
	\textbf{End Function}
\end{algorithm}

In this subsection, we present the design of an adversarial training scheme that is used to optimize the pyramid style extractor $E$ and generator $G$ of COutfitGAN.
For clarity, the entire training process of COutfitGAN is summarized in \textbf{Algorithm}~\ref{algo}.
We first initialize the parameters $\theta_{E}$, $\theta_{G}$, $\theta_D$, and $\theta_{D_{coll}}$ of the pyramid style extractor $E$, outfit generator $G$, real/fake discriminator $D$, and collocation discriminator $D_{coll}$, respectively.
We then sample a batch of data including sample outfits and their corresponding silhouette masks for each fashion item (shown in line 3).
We also sample a batch of masks for masking some fashion items in an outfit to divide the given fashion items and targeted fashion items (shown in line 4).
In the subsequent steps, first the real/fake discriminator is trained (lines 5--8), then the collocation discriminator is trained (lines 10--13).
The style extractor and outfit generator are trained simultaneously (lines 14--18).
For training the real/fake discriminator, the parameters of the other modules are frozen.
After this, fashion items are generated by the style extractor and outfit generator (shown in line 6 and given in detail in lines 20--25).
Then the objective function for the real/fake discriminator is calculated using Eq.~(\ref{l_gan}).
The parameters $\theta_D$ are updated with the gradient of $\mathcal{L}_{gan}$ with a learning rate $\eta$ (lines 8--9).
For training the collocation discriminator $D_{coll}$, the parameters of the other modules are frozen.
First, the collocation loss $\mathcal{L}_{coll_{dis}}$ is calculated using Eq.~(\ref{l_coll_dis}).
The collocation discriminator $D_{coll}$ is optimized by the gradient of $\mathcal{L}_{coll_{dis}}$ with a learning rate $\eta$ (lines 12--13).
Last, the style extractor $E$ and outfit generator $G$ are trained simultaneously, with the parameters of the other modules frozen.
The generation loss $\mathcal{L}_g$ is calculated.
The parameters $\theta_{E}$ and $\theta_{G}$ are updated by the gradient of $\mathcal{L}_g$ with the same learning rate $\eta$ (lines 16--18).
When the training has converged, our framework returns the pyramid style extractor and outfit generator to synthesize complementary fashion items based on the given ones.

\section{Experiments}
In this section, we first describe the construction of our dataset in detail.
The experimental setup and parameter settings are then described.
The performance of our proposed COutfitGAN is compared with that of several competitive image-to-image translation baselines.
Finally, we provide the results of an ablation study to verify the effectiveness of the main modules in COutfitGAN.
\label{experiment}
\subsection{The Dataset}
\begin{figure}[t]
	\centering
	\includegraphics[width=0.45\textwidth, height=0.35857143\textwidth]{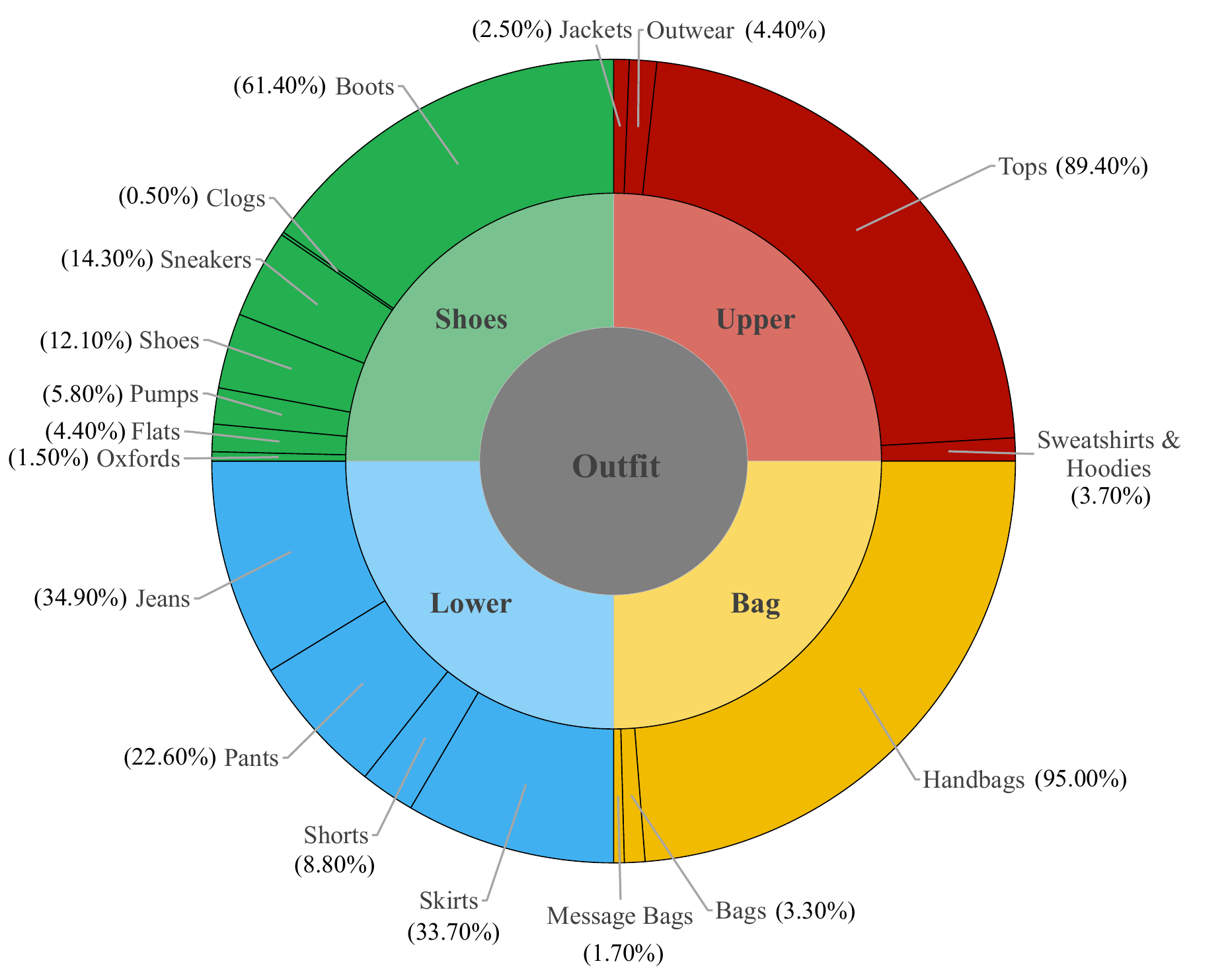}
	\caption{Category distribution of OutfitSet}
	\label{dataset_distr}
	\vspace{-0.5cm}
\end{figure}
Although there are several datasets for fashion compatibility learning, such as UT Zappos50K \cite{yu2014fine}, Maryland Polyvore dataset\cite{han2017learning}, FashionVC \cite{fashionvc}, and IQON3000 \cite{10.1145/3343031.3350956}, which are available for fashion recommendations, all of these datasets lack explicit category annotations for the fashion items in an outfit.
To overcome these limitations in the current datasets and to examine the effectiveness of our proposed outfit generation framework, we collected fashion outfits from a fashion matching community, Polyvore.com, which includes numerous compatible fashion outfits constructed by fashion experts.
The fashion items were assembled into various outfits based on the preferences of fashion experts, with the aim of clearly and attractively presenting specific fashion styles.
The original collected dataset included 344,560 outfits in total, which were composed of 2,131,607 fashion items annotated with their categories by Polyvore.
We selected four types of fashion items as common types for dressing in human daily life: upper clothing, bag, lower clothing, and shoes, which means that $N=4$ in an outfit $\mathcal{O}$.
We therefore kept those outfits that were included in all four categories.
As shown in Fig.~\ref{dataset_distr}, the collected fashion items contain diverse categories associated with fine sub-categories.
Since images of shoes usually have diverse orientations due to different shooting angles, we filtered out images that contained only one shoe, and flipped all the left-oriented shoes horizontally to form right-oriented shoes.
After this preprocessing, 32,043 outfits were retained in the dataset.
Each outfit was also associated with the number of likes upvoted by Polyvore users.
We kept the top 20,000 outfits with the most likes to form our OutfitSet.
We then partitioned these outfits randomly into two folds to form a training set containing 16,000 outfits (80\% of OutfitSet) and a test set consisting of 4,000 outfits (20\% of OutfitSet).
We employed a saliency detector \cite{Liu_2019_CVPR} to detect the silhouette masks of the fashion items.
These silhouettes are used to guide the outfit generation process as one kind of supervision information.
Each silhouette of the fashion item was formulated as $[m]^{256\times 256}$ ($m\in \{0,1\}$), in which a value of one denotes the area of the fashion item and zero denotes the background of the image.
To mimic an outfit generation process with arbitrary given items, we implemented three different settings: one fashion item is given, two fashion items are given, and three fashion items are given.
The categories of these given fashion items are not fixed.
This means that learning a model requires the ability to synthesize complementary items when given any possible combination of fashion items.
To validate the effectiveness of our model, we also sampled the same numbers of outfits to mask some fashion items to construct the settings of given fashion items to validate our fashion outfit generation framework.
Each possible combination of masked fashion items is uniform.

\subsection{Experimental Setup and Parameter Settings}
\label{exp_setting}
In the experiments, all images were resized to $256 \times 256$.
In the training phase, the batch size was set to four, and the number of training iterations for the model was set to 120,000.
All of the experiments were performed on an NVIDIA RTX A6000 graphics card, and the implementation was carried out in PyTorch \cite{paszke2017automatic}.
COutfitGAN was trained with an Adam\cite{kingma2014adam} optimizer with $\beta_1=0$ and $\beta_2=0.99$ and a learning rate of $0.002$.
We also applied R1 regularization\cite{mescheder2018training} on the real/fake discriminator for every 16 iterations, as in \cite{karras2020analyzing}, for training stability.
For the settings of the hyper-parameters in Eq.~(\ref{total_loss}), we set $\lambda_1=100$, $\lambda_2=10$, and $\lambda_3=10$.
\subsection{Evaluation Metrics}
To evaluate the performance of our proposed model, we used a variety of evaluation metrics from three perspectives, which are described in the following.
\subsubsection{Similarity Measurement} A similarity measurement is often used to measure the similarity between a synthesized image and a target image.
We adopted two metrics: the structural similarity index measure (SSIM) \cite{wang2004image} and learned perceptual image patch similarity (LPIPS) \cite{zhang2018unreasonable}.
SSIM is a traditional, widely used image quality index for image comparison.
A higher score indicates a higher similarity.
LPIPS is another common metric used to evaluate the similarity of two images, particularly for a synthesized image and a target image, with a pre-trained deep model.
We used the default pre-trained AlexNet \cite{alexnet} provided by the authors \cite{zhang2018unreasonable} to calculate the LPIPS metric.
Here, a higher score indicates a lower similarity.
\subsubsection{Authenticity Measurement} Previous research \cite{choi2020starganv2} has suggested that the Fr\'{e}chet inception distance (FID)\cite{heusel2017gans} can be used to estimate the authenticity of synthesized images in the feature space mapped through a pre-trained deep model.
More specifically, the FID measures the similarity between two domains of images and is particularly suitable for real images and images synthesized by GANs.
A lower FID score indicates a higher visual authenticity for the synthesized images.
We use the implementation\footnote{\url{https://github.com/mseitzer/pytorch-fid}} of FID to evaluate the performance measurement in terms of visual authenticity.
\subsubsection{Compatibility Measurement} A compatibility measurement is used to gauge the degree of matching between the synthesized outfits.
Fill-in-the-blank (FITB) was proposed by \cite{han2017learning} for the fashion compatibility prediction task, and it has been widely used in other fashion compatibility methods \cite{vasileva2018learning,cui2019dressing, li2020hierarchical, feng2019interpretable,li2017mining}.
To perform a fair evaluation in terms of the compatibility of each outfit, we developed a new metric, called the fashion fill-in-the-blank best times ($\rm{F^2BT}$), based on FITB.
For this metric, we used the open-source toolbox MMFashion\footnote{\url{https://github.com/open-mmlab/mmfashion}} to evaluate the fashion compatibility between the fashion items making up an outfit.
The fashion compatibility prediction module of MMFashion was developed on the basis of the work in \cite{vasileva2018learning} on fashion compatibility prediction.
To enable a fair comparison, the fashion compatibility predictor $\psi$ was trained on the Maryland Polyvore dataset \cite{han2017learning}, meaning that its training set was different from our OutfitSet, and the pre-trained model was provided by MMFashion.
We extend this evaluation metric to outfit generation by counting the number of times that a method beats its counterpart methods.
Formally, it is described as follows:
\begin{equation}
	\rm{F^2BT} = \sum_{i=1}^{N_{\widetilde{\mathcal{O}}}} \prod_{\rm{CM}\in \rm{CMS}} \mathbbm{1} [Comp(\rm{M}_i) > Comp(\rm{CM}_i)],
\end{equation}
where $\rm{CMS}$ denotes all counterpart methods of method $\rm{M}$, $\prod$ denotes the operation of multiplication of all terms, and $\rm{M}_i$ and $\rm{CM}_i$ denote the $i$-th synthesized outfit of $\rm{M}$ and $\rm{CM}$, based on the same given fashion items setting, respectively; $\mathbbm{1}(\cdot)$ is a condition operation that gives one when the input is true or zero when the input is false; and $Comp$ denotes the fashion compatibility score predicted by the pre-trained model $\psi$.
A higher $\rm{F^2BT}$ score indicates a higher compatibility for the synthesized images.
\subsection{Performance Comparison} 
\subsubsection{Compared Methods}
\label{cmp_methods}
Because the outfit generation task takes images as input and outputs a set of images, we compared our method with general image-to-image translation models.
Five state-of-the-art methods are compared with our method: Pix2Pix \cite{pix2pix2017}, Pix2PixHD \cite{wang2018pix2pixHD}, CycleGAN \cite{CycleGAN2017}, MUNIT \cite{huang2018munit}, and CUT \cite{park2020cut}.
These models include both supervised and unsupervised methods.
To make the present paper self-contained, we give a brief introduction to the baselines mentioned above.

\textbf{Pix2Pix} \cite{pix2pix2017} was the first framework to accomplish supervised image-to-image translation.
It uses a GAN loss function and an L1 (or L2) loss function to learn a conditional distribution for supervised problems.

\textbf{Pix2PixHD} \cite{wang2018pix2pixHD} is an improved version of Pix2Pix.
It designs a coarse-to-fine architecture to improve the resolution of the synthesized images.

\textbf{CycleGAN} \cite{CycleGAN2017} was the first framework to accomplish unsupervised image-to-image translation.
It includes two separate generators and two separate discriminators.
It was the first to introduce a cycle reconstruction loss function to constrain a domain-to-domain mapping to be consistent.

\textbf{MUNIT} \cite{huang2018munit} assumes that the images can be disentangled into content code and style code in latent space.
The style code is sampled from a Gaussian distribution to synthesize diverse images in the target domain during the inference phase.

\textbf{CUT} \cite{park2020cut} is based on CycleGAN\cite{CycleGAN2017} and aggregated with self-supervised learning.
It adopts a patch correspondence loss function to improve the performance of unsupervised image-to-image translation.

Note that for our outfit generation, the inputs are concatenated together in the channel dimension and masked if the corresponding fashion item is not given.
The implementations of these models were all based on original codes released by the authors, and the default settings were used for all of the hyper-parameters, as reported in the original papers.

\subsubsection{Comparison of Results}
\begin{figure*}[t]
    \centering
\includegraphics[width=0.68\textwidth, height=0.76086486\textwidth]{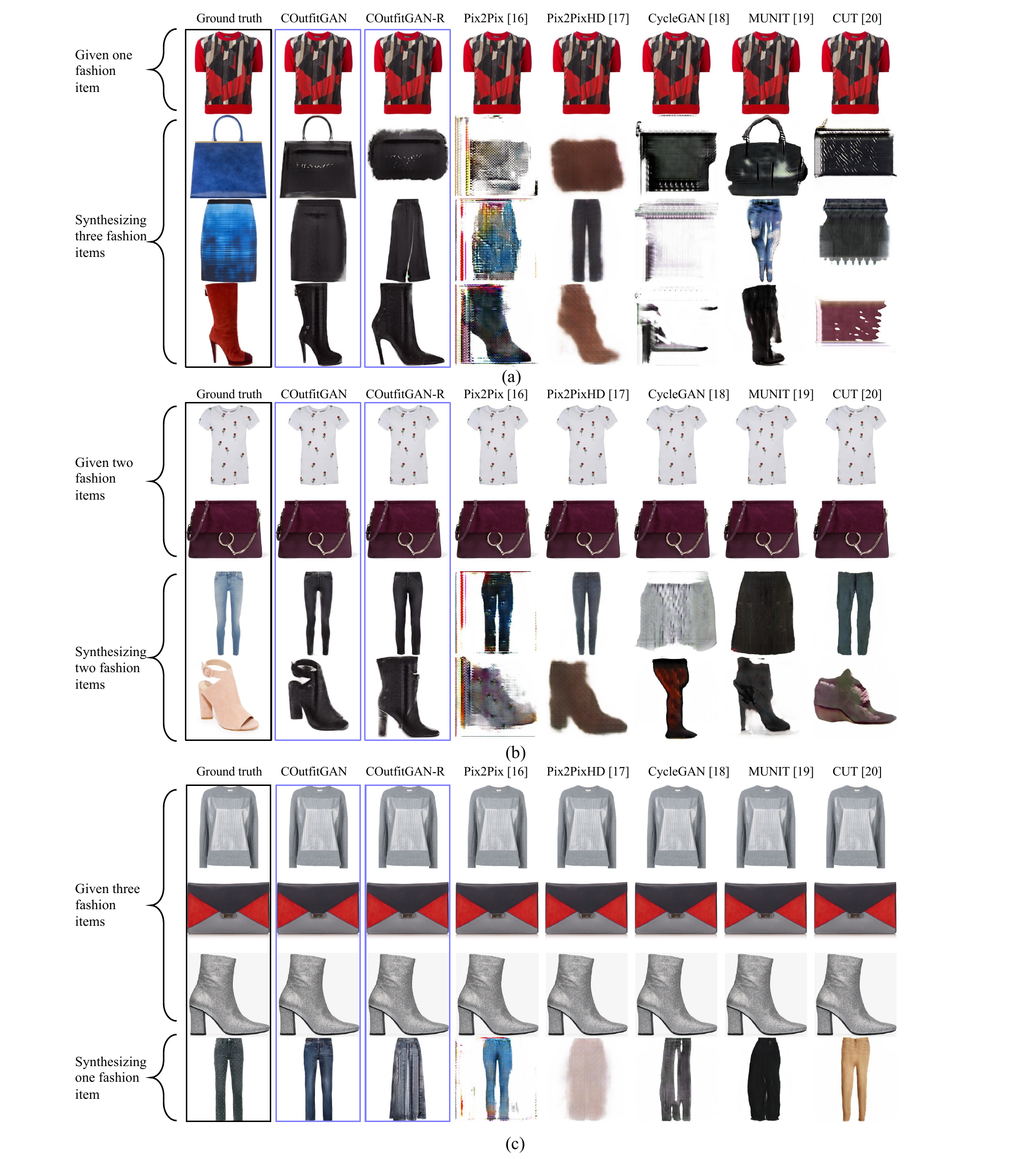}
    \caption{Synthesized samples from COutfitGAN, COutfitGAN-R and other baselines (zoomed in for a better view): (a) synthetic results based on one given fashion item, (b) synthetic results based on two given fashion items, and (c) synthetic results based on three given fashion items.}
    \label{samples}
    \vspace{-0.5cm}
\end{figure*}

A quantitative comparison of the results in terms of all of the evaluation metrics is summarized in Tables \ref{sim_tab}, \ref{aut_tab}, \ref{cmp_tab1}, and \ref{cmp_tab2}.
As described in Section \ref{pro_forl}, silhouette masks that represent the outlines of the target fashion items play an important role in guiding our model.
These reference masks can be divided into two types based on their sources: those provided by a user or those randomly selected by a system.
As shown in Tables \ref{sim_tab}, \ref{aut_tab}, \ref{cmp_tab1}, and \ref{cmp_tab2}, we use COutfitGAN to refer to a version of our model that uses real silhouette masks achieved by saliency detection\cite{Liu_2019_CVPR}, while COutfitGAN-R refers to a version that uses silhouette masks from random selection.
It is worth noting that since different silhouette masks can produce different fashion items, we compare COutfitGAN-R with the baselines only in terms of the authenticity and compatibility measures, i.e., FID and $\rm{F^2BT}$.
Table \ref{sim_tab} shows that our proposed COutfitGAN consistently outperforms other image-to-image translation methods in terms of similarity metrics (SSIM and LPIPS).
Tables \ref{aut_tab}, \ref{cmp_tab1}, and \ref{cmp_tab2} show that our COutfitGAN outperforms the other baselines in terms of the authenticity metric (FID), while COutfitGAN-R outperforms them in terms of the compatibility metric ($\rm{F^2BT}$).

From Fig.~\ref{samples}, it is clear that our model produces superior results, particularly, in terms of the textual details and the harmony of the styles with the given fashion items.
As shown in Table \ref{sim_tab}, the results of a quantitative evaluation show that COutfitGAN yields performance improvements of approximately 0.119, 0.128, and 0.135 in the SSIM when given one, two, and three fashion items, respectively,  with 0.127 being their average value, in comparison to the second-best method, Pix2PixHD; it also reduces the LPIPS by approximately 0.164, 0.168, and 0.158 when given one, two, and three fashion items, respectively, with an average reduction of 0.168, in comparison to the second-best method.
This result suggests that our synthesized images can maintain the overall image structure and visual similarity better than other methods
(see also Table \ref{sim_tab}). 
Fig.~\ref{samples} also shows that COutfitGAN can synthesize the most similar results in terms of visual appearance.
This indicates that our approach not only surpasses other methods in terms of the quantitative similarity metrics, but also outperforms them in terms of human visual observation.

\begin{table}[h]\footnotesize
\centering
\caption{Comparison of COutfitGAN with baselines in terms of similarity metrics.}
\label{sim_tab}
\begin{tabular}{p{2cm} P{0.35cm} P{0.35cm} P{0.35cm} P{0.4cm} P{0.001cm} P{0.35cm} P{0.35cm} P{0.35cm} P{0.4cm}}
\toprule
\multirow{2}{*}{Method} & \multicolumn{4}{c}{SSIM($\uparrow$)} & & \multicolumn{4}{c}{LPIPS($\downarrow$)}\\
& 1&2 &3 &Avg.& & 1&2&3&Avg.\\
\cline{1-6} \cline{7-10} \\
Pix2Pix\cite{pix2pix2017} & 0.422 & 0.396 & 0.366 & 0.395 & &0.545 & 0.557 & 0.575 & 0.559\\
Pix2PixHD\cite{wang2018pix2pixHD} & 0.548 & 0.544 & 0.545 &0.546 & & 0.488 & 0.488 & 0.486 & 0.488\\
CycleGAN\cite{CycleGAN2017} & 0.420 & 0.425 & 0.512 & 0.452 & & 0.573 & 0.541 & 0.485 & 0.533\\
MUNIT\cite{huang2018munit} & 0.443 & 0.478 & 0.496 & 0.472 && 0.530 & 0.510 & 0.503 & 0.514\\
CUT\cite{park2020cut} & 0.473  & 0.489 & 0.529 & 0.497 && 0.534 & 0.513 & 0.473 & 0.506 \\
\hline
COutfitGAN & \textbf{0.667} & \textbf{0.672} & \textbf{0.680} & \textbf{0.673} && \textbf{0.324} & \textbf{0.320} & \textbf{0.315} & \textbf{0.320}\\
\bottomrule
\end{tabular}
\vspace{-0.2cm}
\end{table}

We also compared our models with baseline methods in terms of the authenticity measurement, i.e., the FID.
For each setting of the number of given fashion items, we evaluated the FID of the synthesized and target fashion items.
As shown in Table \ref{aut_tab}, COutfitGAN reduces the FID by approximately 93.6, 39.1, and 37.3 when given one, two, and three fashion items, respectively, with an average reduction by 69.5, in comparison with the second-best method.
Our model with randomly selected silhouette masks, COutfitGAN-R, reduces the FID by approximately 92.6, 38.6, and 37.3, when given one, two, and three fashion items, respectively, for an average reduction of 69.5, in comparison with the second-best method.
From the synthesized results illustrated in Fig.~\ref{samples}, we can see that the images produced by our models have greater authenticity based on human observation.
In particular, the Pix2Pix method sometimes produces checkerboard artifacts, and its synthesized images are not well contoured.
Pix2PixHD always produces blurry images without fine details.
We attribute this to the fact that it may be caused by the coarse-to-fine manner for synthesizing images in Pix2PixHD.
CycleGAN always translates the given images to target images while retaining weird outlines.
This can be attributed to the cycle reconstruction loss function.
Furthermore, the synthesized results from MUNIT and CUT are blurred and not well contoured.
In contrast, our COutfitGAN is able to synthesize the most visually plausible results based on real silhouette masks.
Using randomly selected silhouette masks, our COutfitGAN-R can also synthesize plausible results.
\begin{table}[h]
\centering
\caption{Comparison of COutfitGAN and COutfitGAN-R with baselines in terms of authenticity metrics.}
\label{aut_tab}
\begin{tabular}{p{2cm} P{0.5cm} P{0.5cm} P{0.5cm} P{0.7cm}}
\toprule
\multirow{2}{*}{Method} & \multicolumn{4}{c}{FID($\downarrow$)}\\
&1&2&3 &Avg.\\
\cline{1-5} \\
Pix2Pix\cite{pix2pix2017} & 191.4  & 185.9 & 203.8 & 193.7 \\
Pix2PixHD\cite{wang2018pix2pixHD} & 246.2 & 249.7 & 254.4 & 250.1  \\
CycleGAN\cite{CycleGAN2017} & 249.9 & 183.2 & 172.3 & 201.8\\
MUNIT\cite{huang2018munit} & 144.1 & 109.5 & 114.0 & 122.5 \\
CUT\cite{park2020cut} & 216.0 & 91.7 & 93.4 & 133.7 \\
\hline
COutfitGAN & \textbf{50.5} & \textbf{52.6} & \textbf{56.1} & \textbf{53.1}\\
COutfitGAN-R & \textbf{51.1} & \textbf{53.1} & \textbf{56.1} & \textbf{53.1}\\
\bottomrule
\end{tabular}
\end{table}

With respect to the compatibility measurement for the synthesized outfits, the results for $\rm{F^2BT}$ for our COutfitGAN and COutfitGAN-R suggest that a generator supervised by our collocation discriminator can produce synthetic outfits with a superior degree of matching in comparison to other baselines, as shown in Tables \ref{cmp_tab1} and \ref{cmp_tab2}.
Pix2PixHD also produces promising results for the $\rm{F^2BT}$, as shown in Tables \ref{cmp_tab1} and \ref{cmp_tab2}; however, the outfits synthesized by Pix2PixHD are always similar and blurred, as shown in Fig.~\ref{samples}.
The outfits synthesized by Pix2PixHD therefore did not achieve a high compatibility score from a human perspective, due to the lack of fine details and aesthetics.
\begin{table}[h]
\centering
\caption{Comparison of COutfitGAN with baselines in terms of compatibility metrics.}
\label{cmp_tab1}
\begin{tabular}{p{2cm} P{0.5cm} P{0.5cm} P{0.5cm} P{0.56cm}}
\toprule
\multirow{2}{*}{Method} & \multicolumn{4}{c}{$\rm{F^2BT}$($\uparrow$)}\\
&1&2&3&Avg.\\
\cline{1-5} \\
Pix2Pix\cite{pix2pix2017} & 569 & 473 & 531 & 789\\
Pix2PixHD\cite{wang2018pix2pixHD} & 1535 & 1328 & 971 & 1278\\
CycleGAN\cite{CycleGAN2017} &  322 & 519 & 1044 & 628\\
MUNIT\cite{huang2018munit} & 391 & 473 & 531 & 465\\
CUT\cite{park2020cut} & 542 & 1089 & 1043 & 891\\
\hline
COutfitGAN & \textbf{2641} & \textbf{1768} & \textbf{1436} & \textbf{1948}\\
\bottomrule
\end{tabular}
\vspace{-0.2cm}
\end{table}

\begin{table}[h]
\centering
\caption{Comparison of COutfitGAN-R with baselines in terms of compatibility metrics.}
\label{cmp_tab2}
\begin{tabular}{p{2cm} P{0.5cm} P{0.5cm} P{0.5cm} P{0.56cm}}
\toprule
\multirow{2}{*}{Method} & \multicolumn{4}{c}{$\rm{F^2BT}$($\uparrow$)}\\
&1&2&3&Avg.\\
\cline{1-5} \\
Pix2Pix\cite{pix2pix2017} & 576 & 861 & 930 &789\\
Pix2PixHD\cite{wang2018pix2pixHD} & 1551 & 1287 & 1037 &1292\\
CycleGAN\cite{CycleGAN2017} & 339 & 557 & 1028 & 641 \\
MUNIT\cite{huang2018munit} & 385 & 460 & 556 & 467\\
CUT\cite{park2020cut} & 528 & 1102 & 1078 & 903\\
\hline
COutfitGAN-R & \textbf{2621} & \textbf{1733} & \textbf{1371} & \textbf{1903}\\
\bottomrule
\end{tabular}
\vspace{-0.2cm}
\end{table}

\subsection{Ablation Study}
This subsection presents the results of four sets of ablation studies carried out to validate the effectiveness of the silhouette masks, the pyramid style extractor, the UNet-based real/fake discriminator, and the collocation discriminator, which are the main components of COutfitGAN.

\subsubsection{Effectiveness of the Silhouette Masks} To investigate the effectiveness of silhouette masks, we validate it in two aspects.
First, we modified the compared methods mentioned in Section \ref{cmp_methods}. The masked fashion items of input were replaced with their real silhouette masks. We compared the original image-to-image baselines with the modified versions. The comparative results are summarized in Table \ref{baselines_with_masks}. As shown in Table \ref{baselines_with_masks}, taking silhouette masks as reference information can indeed improve the performance of general image-to-image translation methods, except for MUNIT, in terms of the metrics of SSIM, LPIPS, and FID, when the inputs and outputs have no explicit spatial mapping. For MUNIT, we find that it has not gained improvements even when the silhouette masks are given. This may be ascribed to the fact that MUNIT tries to disentangle images into content and style codes, which are expected to fit a Gaussian distribution, and then synthesizes diverse images with noises sampled from this known distribution. However, content and style codes are represented as vectors, while the silhouette masks cannot be represented as vectors to preserve their spatial information well. On the other hand, as a relative comparison metric, $\rm{F^2BT}$ of our COutfitGAN changes from 1948 to 1700. This indicates that the compared methods with silhouette masks have improved their compatibility metrics. Here, it should be noted that, although most of the modified methods have performance improvements, our proposed COutfitGAN still achieves the best performance over all compared methods. Second, to validate the effectiveness of silhouette masks in our proposed COutfitGAN, we conducted an ablation study by removing the silhouette masks of all silhouette and style (S--S) blocks in our outfit generator. However, according to our implementation, COutfitGAN without silhouette masks cannot converge and can only synthesize images with background colors. The most likely cause of the non-convergence phenomenon may be ascribed to the fact that only taking vector-based embeddings as inputs cannot assure our model to learn the complex mapping function from the given fashion items to the targeted ones. In particular, if the silhouette masks are not given, the outfit generator of COutfitGAN synthesizes the targeted fashion items by only taking the style embeddings as input. Specifically, these embeddings are composed of 14 different 512-dimensional vectors. The dimension size of the vectors is much smaller than that of the synthesized images, and their vector-based representations cannot preserve the spatial information of the targeted fashion items well. This indicates that a silhouette mask, as a kind of information compensation, plays an important role in guaranteeing the performance of our COutfitGAN. Meanwhile, it should be noted that, although our model without using silhouette masks produces worse results in comparison with the baselines without using silhouette masks, these masks can be easily obtained in practice to enable our COutfitGAN to produce satisfactory results on our task.

\begin{table}[h]
\centering
\caption{Comparisons of COutfitGAN with the baselines using silhouette masks (here, all results are the average results from three different settings of the number of given fashion items).}
\label{baselines_with_masks}
\begin{tabular}{p{3.6cm} P{0.85cm} P{0.85cm} P{0.85cm} P{0.85cm}}
\toprule
Method & SSIM($\uparrow$) & LPIPS($\downarrow$) & FID($\downarrow$) & $\rm{F^2BT}$($\uparrow$)\\
\hline
Pix2Pix (+ silhouette mask) & 0.623 & 0.372 & 118.02 & 894 \\
Pix2PixHD (+ silhouette mask) & 0.639 & 0.357 & 122.77 & 889\\
CycleGAN (+ silhouette mask) & 0.559 & 0.415 & 114.60 & 725\\
MUNIT (+ silhouette mask) & 0.348 & 0.664 & 263.99 & 812\\
CUT (+ silhouette mask) & 0.585 & 0.371 & 66.94 & 979\\
\hline
COutfitGAN (ours)& \textbf{0.673} & \textbf{0.320} &\textbf{53.06} & \textbf{1700}\\
\bottomrule
\end{tabular}
\vspace{-0.2cm}
\end{table}

\subsubsection{Effectiveness of the Pyramid Style Extractor} 
\begin{figure}[t]
	\centering
	\includegraphics[width=0.5\textwidth, height=0.3189\textwidth]{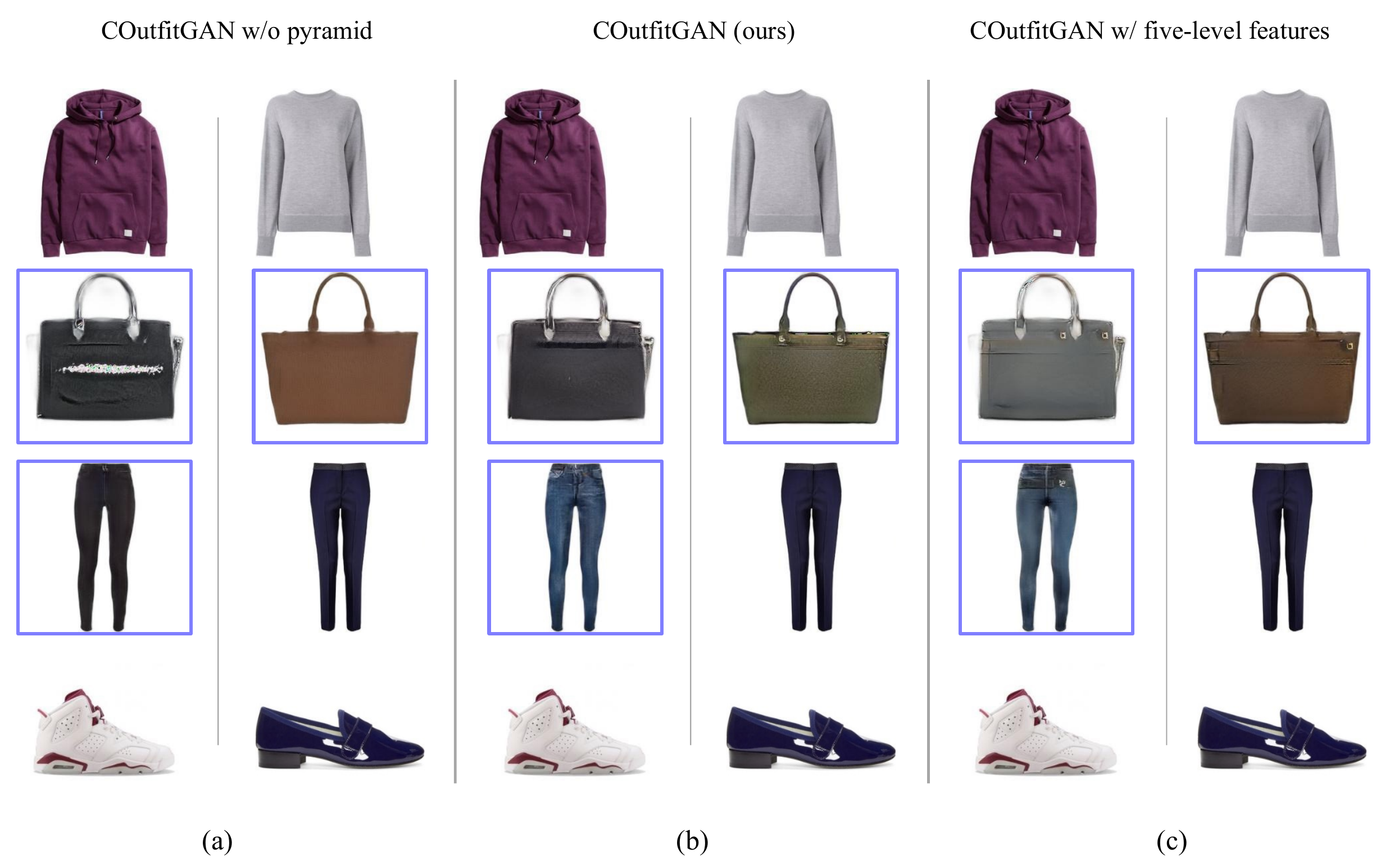}
	\caption{Synthesized samples of COutfitGAN with respect to pyramid style extractor (here, violet rectangles represent the synthetic fashion items): (a) COutfitGAN without pyramid style extractor, (b) COutfitGAN (ours), and (c) COutfitGAN with five-level features.}
	\label{pyramid_vs_no_pyramid}
	\vspace{-0.5cm}
\end{figure}

To investigate the effectiveness of the pyramid style extractor, we validate it from three perspectives.
First, we trained COutfitGAN without using the pyramid style extractor. Then we compared this version with our original COutfitGAN in terms of the aforementioned four metrics.
In Table \ref{pse_nopse}, `COutfitGAN w/o pyramid' means that we trained the COutfitGAN without pyramid features and only used the last features extracted by the pyramid style extractor.
The results show that with a pyramid style extractor, our COutfitGAN produces a similar result in terms of SSIM and a slight decrease in terms of LPIPS. Meanwhile, our COutfitGAN produces better results in terms of authenticity and compatibility measurements.
Especially for the compatibility metric, the pyramid style extractor significantly improves the compatibility of the synthesized outfits. We also observe that the pyramid style extractor concentrates more on the finer texture as shown in Fig.~\ref{pyramid_vs_no_pyramid}. The pyramid architecture is able to synthesize better texture as reported by \cite{heeger1995pyramid}. In addition, the texture is a key factor influencing the compatibility of an outfit\cite{kim2021self}. The decrease in similarity metrics may be ascribed to the fact that the pyramid style extractor concentrates more on texture generation rather than the overall similarity between synthesized results and the ground-truth images. In contrast, focusing too much on similarity may lead to synthesizing blurry images that lack details. To validate whether more pyramid features improve the performance continuously or not, we also conducted an experiment to improve visual embeddings from three-level to five-level features. However, we find that our COutfitGAN with three-level features and COutfitGAN with five-level features achieve similar results in terms of all metrics as shown in Table \ref{pse_nopse}. In addition, the COutfitGAN with five-level features needs more training iterations to converge in comparison with our proposed COutfitGAN. For clarity, we also provide some sampled synthetic results of these two settings of COutfitGAN as shown in Fig.~\ref{pyramid_vs_no_pyramid}. We observe that these two versions of COutfitGAN produce similar results. Thus, three-level features are sufficient for our COutfitGAN to synthesize photo-realistic and compatible fashion items based on the given ones.
\begin{table}[h]
\centering
\caption{Comparisons of COutfitGAN with and without the pyramid style extractor (here, all results are the average results from three different settings of the number of given fashion items).}
\label{pse_nopse}
\begin{tabular}{p{3.9cm} P{0.75cm} P{0.75cm} P{0.75cm} P{0.75cm}}
\toprule
Method & SSIM($\uparrow$) & LPIPS($\downarrow$) & FID($\downarrow$) & $\rm{F^2BT}$($\uparrow$)\\
\hline
COutfitGAN w/o pyramid & \textbf{0.676} & \textbf{0.306} & 55.49 & 1787\\
COutfitGAN (ours) & 0.673 & 0.320 & \textbf{53.06} & \textbf{1948}\\
COutfitGAN w/ five-level features & \textbf{0.676} & 0.323 & 54.08 & 1905\\
\bottomrule
\end{tabular}
\end{table}

\subsubsection{Effectiveness of the UNet-based Real/Fake Discriminator}
We also explore the impact of the UNet-based real/fake discriminator in COutfitGAN on its performance.
More specifically, we examined its effectiveness on the FID in terms of visual authenticity.
The results of the comparisons are given in Table \ref{unet_nounet}, where `COutfitGAN w/o UNet discriminator' refers to our model without using the UNet-based discriminator, and with the default discriminator in StyleGAN2.
We can observe that the model without the UNet-based real/fake discriminator has lower FID values, which are reduced from 60.5, 60.2, and 61.04 to 50.5, 52.6, and 56.1, when given one, two, and three fashion items, respectively, with an average reduction from 60.7 to 53.1.
These quantitative results suggest that our UNet-based real/fake discriminator can effectively improve the visual authenticity of the synthesized fashion items.
\begin{table}[h]
\centering
\caption{Comparison of COutfitGAN with and without the UNet-based discriminator in terms of authenticity metrics.}
\label{unet_nounet}
\begin{tabular}{p{4.2cm} P{0.5cm} P{0.5cm} P{0.5cm} P{0.5cm}}
\toprule
\multirow{2}{*}{Method} & \multicolumn{4}{c}{FID($\downarrow$)}\\
&1&2&3 &Avg.\\
\cline{1-5} \\
COutfitGAN w/o UNet discriminator & 60.5 & 60.2 & 61.4 & 60.7\\
\hline
COutfitGAN (ours)& \textbf{50.5} & \textbf{52.6} & \textbf{56.1} & \textbf{53.1}\\
\bottomrule
\end{tabular}
\end{table}
\subsubsection{Effectiveness of the Collocation Discriminator}
\begin{figure}[t]
	\centering
	\includegraphics[width=0.49\textwidth, height=0.25852761\textwidth]{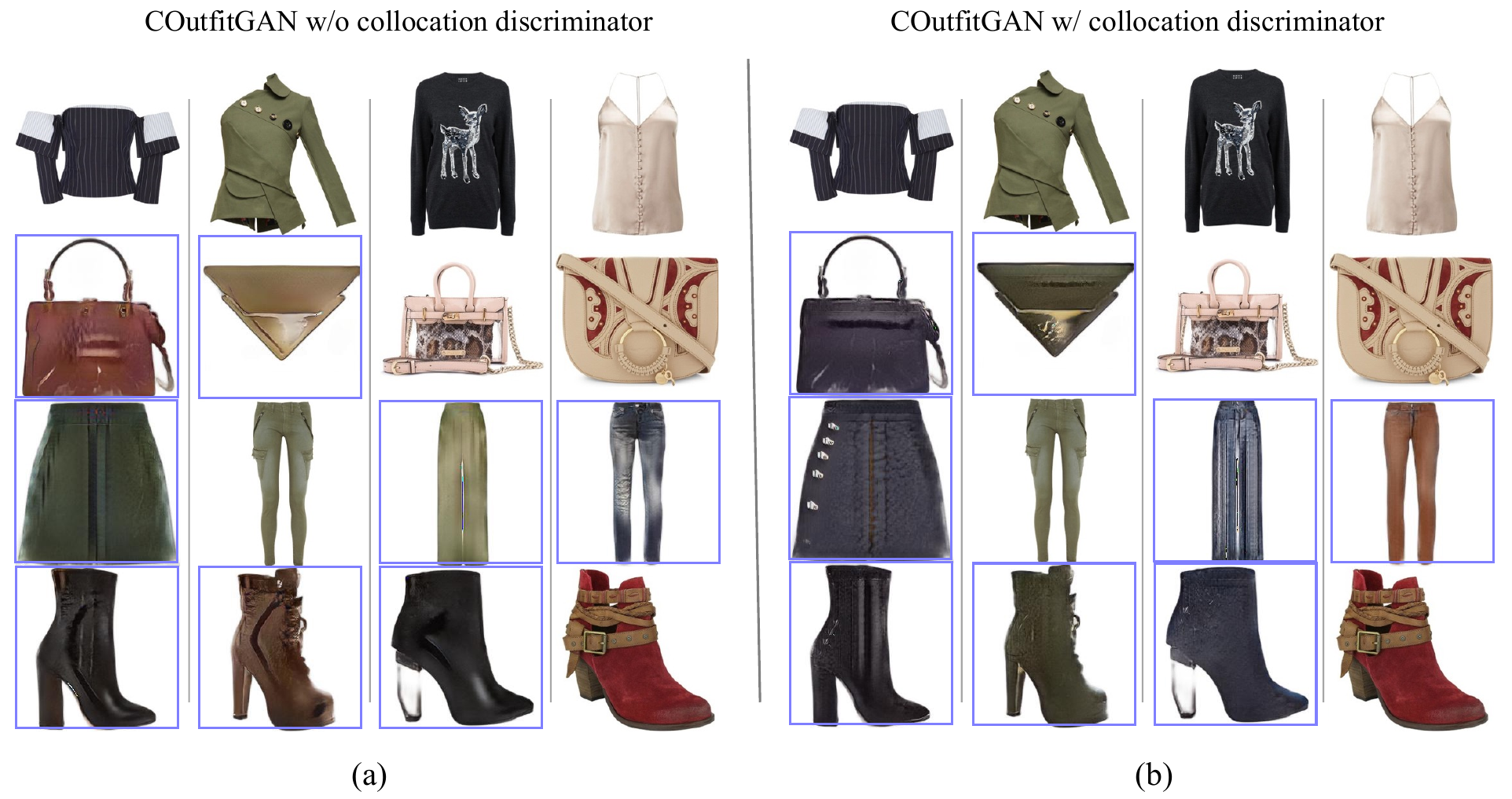}
	\caption{Synthesized samples of COutfitGAN with respect to the collocation discriminator module (here, violet rectangles represent the synthetic fashion items): (a) COutfitGAN without the collocation discriminator and (b) COutfitGAN with the collocation discriminator.}
	\label{cmp_no_cmp_samples}
	\vspace{-0.5cm}
\end{figure}

In addition to the two key components of COutfitGAN examined above, we further investigated the impact of our proposed collocation discriminator on the performance in terms of fashion compatibility.
The results of the comparison are given in Table \ref{cd_no_cd}, where `COutfitGAN w/o collocation discriminator' refers to our model trained without a collocation discriminator.
We can see that the model without a collocation discriminator suffers a significant decrease in $\rm{F^2BT}$ from 2641, 1768, and 1436 to 1299, 1204, and 1149, when given one, two, and three fashion items, respectively, with an average reduction from 1948 to 1217.
In addition, Fig.~\ref{cmp_no_cmp_samples} shows that COutfitGAN with the developed collocation discriminator synthesizes more compatible outfits with more harmonious styles than the model without the collocation discriminator.
The collocation module enhances the co-occurrence frequency of compatible elements or styles for compatible outfits.
These quantitative and qualitative results suggest that the collocation discriminator can effectively improve the compatibility of the synthesized outfits.

\begin{table}[h]
	\centering
	\caption{Comparison of COutfitGAN with and without collocation discriminator in terms of fashion compatibility metrics.}
	\label{cd_no_cd}
	\begin{tabular}{p{4.9cm} P{0.48cm} P{0.48cm} P{0.48cm} P{0.54cm}}
		\toprule
		\multirow{2}{*}{Method} & \multicolumn{4}{c}{$\rm{F^2BT}$($\uparrow$)}\\
		&1&2&3&Avg.\\
		\hline \\
		COutfitGAN w/o collocation discriminator & 1299 & 1204 & 1149 & 1217\\
		\hline
		COutfitGAN (ours) & \textbf{2641} & \textbf{1768} & \textbf{1436} & \textbf{1948}\\
		\bottomrule
\end{tabular}
\vspace{-0.2cm}
\end{table}

\section{Conclusion}

This research has presented an outfit generation framework, COutfitGAN, with the aim of synthesizing photo-realistic images of fashion items that are compatible with arbitrarily given fashion items.
In particular, to exploit the styles of the given fashion items shared in a compatible outfit, COutfitGAN uses a silhouette and style fusion strategy for image synthesis which can overcome the spatial misalignment issue that arises in general image-to-image translation methods.
COutfitGAN includes a pyramid style extractor, an outfit generator, a UNet-based real/fake discriminator, and a collocation discriminator.
The pyramid style extractor adopts a sequential and pyramid architecture to extract the styles of the given fashion items to be used in the style learning for the target fashion items.
Then our outfit generator fuses silhouette masks and styles to synthesize complementary fashion items.
The UNet-based real/fake discriminator is employed to enhance the image quality in terms of visual authenticity.
Our collocation discriminator is built on contrastive learning to maintain a better compatibility of the synthesized outfits.
To evaluate the effectiveness of the proposed model, we constructed a large-scale dataset that consists of 20,000 outfits associated with 80,000 fashion items.
Extensive experimental results show that our method can achieve state-of-the-art performance on the task of outfit generation in comparison to existing methods.
In the future, we plan to concentrate on synthesizing outfits with finer details, and to use no or other reference information, such as textual descriptions, to replace the silhouette masks in a more flexible manner to guide the process of outfit generation.
\label{cnc}
\bibliographystyle{IEEEtran}
\bibliography{ref}


%
%

\vfill

\end{document}